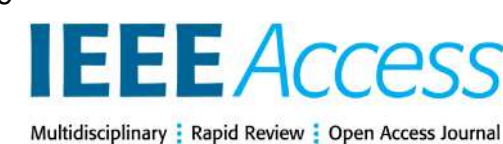

# A Data-Driven Review of Remote Sensing-Based Data Fusion in Precision Agriculture from Foundational to Transformer-Based Techniques

**Mahdi Saki[1], Rasool Keshavarz[1], Senior Member, IEEE, Daniel Franklin[1], Member, IEEE, Mehran Abolhasan[1], Senior Member, IEEE, Justin Lipman[1], Senior Member, IEEE, Negin Shariati[1], Senior Member, IEEE**

[1]RF and Communication Technologies (RFCT) Research Laboratory, University of Technology Sydney, Ultimo, NSW 2007, Australia

Corresponding author: Mahdi Saki (email:Mahdi.Saki@uts.edu.au).

This research was supported by NTT Group and the Food Agility Cooperative Research Centre (CRC) Ltd, funded under the Commonwealth Government CRC Program for the 'Sustainable Sensing, Enhanced Connectivity, and Data Analytics for Precision Urban and Rural Agriculture' project at the RFCT laboratory at the University of Technology Sydney (UTS).

**ABSTRACT** This review explores recent advancements in data fusion techniques and Transformer-based remote sensing applications in precision agriculture. Using a systematic, data-driven approach, we analyse research trends from 1994 to 2024, identifying key developments in data fusion, remote sensing, and AI-driven agricultural monitoring. While traditional machine learning and deep learning approaches have demonstrated effectiveness in agricultural decision-making, challenges such as limited scalability, suboptimal feature extraction, and reliance on extensive labelled data persist. This study examines the comparative advantages of Transformer-based fusion methods, particularly their ability to model spatiotemporal dependencies and integrate heterogeneous datasets for applications in soil analysis, crop classification, yield prediction, and disease detection. A comparative analysis of multimodal data fusion approaches is conducted, evaluating data types, fusion techniques, and remote sensing platforms. We demonstrate how Transformers outperform conventional models by enhancing prediction accuracy, mitigating feature redundancy, and optimizing large-scale data integration. Furthermore, we propose a structured roadmap for implementing data fusion in agricultural remote sensing, outlining best practices for ground-truth data selection, platform integration, and fusion model design. By addressing key research gaps and providing a strategic framework, this review offers valuable insights for advancing precision agriculture through AI-driven data fusion techniques.

## I. INTRODUCTION

In response to global challenges such as environmental sustainability, climate change, and an expanding human population projected to reach nearly 10 billion by 2050, agriculture must evolve to meet increasing demands for food, fibre, fuel, and other raw materials [1], [2]. Precision agriculture (PA) has emerged as a solution to these challenges, utilizing advanced technologies such as remote sensing, global positioning system (GPS), geographic information systems (GIS), and artificial intelligence (AI) to optimize agricultural inputs and improve crop yields [3], [4].

PA is a critical means to enhance the efficiency of agricultural production using advanced information technology (IT) [5]. The implementation of PA, also known as smart farming, relies on the ability to collect, process, and analyse spatial and temporal data to optimize field management practices [5], [6]. Despite its enormous potential, the adoption of PA remains below expectations due to factors such as high initial investment costs, the complexity of IT, and the need for specialized knowledge [5].

Remote sensing (RS) has seen rapid advancements and widespread adoption in PA, offering high-resolution data for applications ranging from crop monitoring to irrigation management [3]. Remote sensing has proven to be an effective tool for capturing and monitoring the spectral and temporal properties of the land surface influenced by human activities at different spatial and temporal scales [7]. The use of satellite imagery, unmanned Aerial vehicles (UAVs), and ground-based sensors [8], [9], [10] as enabled detailed mapping of crop health, soil properties, and environmental conditions, facilitating targeted interventions and resource-efficient farming practices [3].

NOMENCLATURE

| | |
|---|---|
| AI | Artificial Intelligence |
| CNN | Convolutional Neural Network |
| DEM | Digital Elevation Model |
| DF | Data Fusion |
| DL | Deep Learning |
| GPR | Ground Penetrating Radar |
| HS | Hyperspectral |
| HSI | Hyperspectral Imaging |
| INFRD | Infrared |
| IoT | Internet Of Things |
| IT | Information Technology |
| LiDAR | Light Detection and Ranging |
| LSTM | Long Short-Term Memory |
| LULC | Land Use Land Cover |
| ML | Machine Learning |
| MLP | Multilayer Perceptron |
| MSI | Multispectral Imaging |
| NDVI | Normalized Difference Vegetation Index |
| PA | Precision Agriculture |
| RGB | Red, Green and Blue |
| RNN | Recurrent Neural Network |
| RS | Remote Sensing |
| SAR | Synthetic Aperture Radar |
| STARF | Spatial And Temporal Adaptive Reflectance Fusion |
| TFT | Temporal Fusion Transformer |
| UAV | Unmanned Aerial Vehicle |
| ViT | Vision Transformer |

Remote sensors used in agriculture can be broadly categorized based on the technology and techniques they employ. Optical remote sensing systems, which include multi-spectral and hyperspectral imaging, utilize visible, near-infrared, and short-wave infrared sensors to capture detailed images of the Earth's surface [11], [12], [13]. These systems are particularly effective in assessing vegetation health, soil properties, and water content due to their high spatial resolution. However, optical systems are heavily dependent on sunlight and are therefore ineffective at night or in cloudy conditions. Additionally, atmospheric interference, such as haze and cloud cover, can degrade the quality of the collected data, limiting their effectiveness in certain environments.

Radar remote sensing systems, such as Synthetic Aperture Radar (SAR), use microwave radiation to penetrate clouds and operate independently of daylight. This capability makes radar systems particularly valuable in regions with frequent cloud cover or during seasons with limited daylight. Radar systems are also adept at analysing surface structures, soil moisture, and crop conditions, providing insights that optical sensors might miss. Despite these advantages, radar data can be more complex to interpret, requiring advanced processing techniques. Furthermore, radar sensors generally offer lower spatial resolution compared to optical systems, which may limit their effectiveness in detailed crop monitoring [14].

Thermal infrared remote sensing systems detect heat emitted from the Earth's surface, allowing for the study of surface temperatures and thermal properties. These systems are crucial for monitoring plant stress, water usage, and identifying areas of potential heat stress in crops. Thermal sensors have the advantage of functioning effectively both day and night, providing continuous monitoring capabilities. However, they often provide coarser spatial resolution compared to optical systems, potentially missing small-scale variations. Additionally, the cost of thermal imaging technology can be prohibitive, especially for high-resolution applications [15].

LiDAR (Light Detection and Ranging) systems, which measure distances by illuminating the target with laser light and measuring the time it takes for the light to return, are highly effective in generating detailed topographic maps. This capability is essential for terrain analysis, forest structure studies, and precision agriculture. LiDAR systems are particularly advantageous in capturing ground elevation data even in areas with dense vegetation, providing valuable insights into both ground and canopy structures. Despite these benefits, LiDAR systems are expensive to deploy and operate, often requiring specialized equipment and expertise. Furthermore, the area covered by LiDAR in a single scan is typically smaller compared to satellite-based systems, making it less suitable for large-scale agricultural monitoring [16].

RS technologies also provide a diagnostic tool that can serve as an early warning system, allowing the agricultural community to intervene early on to counter potential problems before they spread widely and negatively impact crop productivity. Despite recent advancements in sensor technologies, data management, and data analytics, the agricultural sector is yet to fully implement RS technologies due to knowledge gaps regarding their sufficiency, appropriateness, and techno-economic feasibilities [17].

Data fusion is the process of integrating data from multiple sensors or platforms to enhance the accuracy and utility of remote sensing information. In agriculture, data fusion can occur between various combinations of sensors, each offering unique advantages and challenges [18].

Satellite-to-satellite data fusion involves combining data from satellites with different frequencies, polarizations, or revisit times, enhancing the ability to monitor diverse crop and soil parameters, such as moisture content and vegetation health.

This approach increases temporal resolution, allowing for more frequent monitoring and timely interventions. However, the process of integrating data from different satellites is complex, requiring sophisticated algorithms and advanced technical expertise. Additionally, the cost of accessing and processing data from multiple satellites can be significant, particularly when high-resolution sensors are involved [19].

The fusion of satellite and airborne/drone-based sensors data leverages the complementary strengths of each platform. While satellite data provides broad coverage, airborne data offers high spatial resolution, creating a comprehensive view of the agricultural landscape. Drones can be deployed to verify and investigate specific areas identified by satellite observations, enhancing the precision of monitoring. However, aligning data from these different platforms requires careful synchronization, particularly in terms of timing and spatial resolution. The operational challenges of integrating satellite and drone data, including handling different data formats and processing techniques, can be resource intensive [20].

The fusion of ground sensors with satellite or drone data enhances accuracy by providing precise, localized data that can validate and calibrate satellite or drone observations. This multi-layered analysis enables more informed decision-making in farm management. However, the integration of ground sensors with aerial or satellite data can be technically challenging, requiring advanced algorithms and processing capabilities. Moreover, the deployment and maintenance of ground sensors, along with the acquisition of satellite or drone data, can be costly and resource-intensive, limiting accessibility for some agricultural operations [21].

Therefore, the complexity of integrating diverse data sources in remote sensing demands sophisticated approaches for analysis and interpretation. Machine learning (ML) and artificial intelligence (AI) have emerged as crucial approaches and enabling technologies in this context, automating the process of identifying patterns, anomalies, and trends in agricultural monitoring. AI-driven models can continuously improve their accuracy by adapting to new data inputs, making them highly effective for real-time monitoring and prediction in agriculture. Furthermore, ML models can forecast future agricultural conditions based on historical data, enabling proactive management practices. AI techniques also aid in reducing the complexity of fused data, making it more manageable and interpretable for decision-makers. However, challenges remain, including the need for high-quality data inputs and significant computational resources, which can limit the accessibility of these technologies for smaller agricultural operations [22].

The intersection of AI - particularly deep learning (DL) - with agricultural remote sensing has enabled significant advancements in PA. The increasing complexity of agricultural challenges like population growth, climate change, and the global COVID-19 pandemic necessitates innovative solutions to enhance crop yield [23] while promoting sustainable farming practices. High-performance sensors [24] and autonomous systems [25], integrated with DL techniques, have emerged as critical tools in this context. These technologies facilitate the non-invasive monitoring of crops at a range of spatial and temporal scales, leveraging RGB, multispectral, hyperspectral, thermal, and synthetic aperture radar (SAR) sensors mounted on satellites, UAVs, and terrestrial robots. The massive datasets generated offer rich crop information but also present challenges in data acquisition, processing and interpretation. Deep learning, with its powerful feature-learning capabilities, addresses these challenges by enabling sophisticated analysis of high-resolution images for applications such as land mapping, crop classification, stress monitoring, and yield prediction [26].

Moreover, the rapid advancements in DL particularly the introduction of Transformer models, have significantly impacted the field of remote sensing applications, especially in the areas of very high-resolution (VHR) imagery, hyperspectral imaging (HSI), and SAR [27], [28]. Transformers, originally developed for natural language processing, have demonstrated remarkable success in various computer vision tasks due to their capability to model long-range dependencies and global contextual information. These models have been increasingly applied to remote sensing image analysis, offering substantial improvements in tasks such as land use/land cover (LULC) classification, segmentation, fusion, change detection, and object detection. The shift from convolutional neural networks (CNNs) to Transformer-based architectures addresses the limitations of CNNs, such as their constrained receptive field and inability to fully capture global information [28]. Recent advancements include the application of vision Transformers in various remote sensing tasks such as classification, object detection, and change detection. The ability of Transformers to model complex relationships and adapt to different scales has made them a valuable tool in remote sensing [27].

Recent reviews have explored the role of Transformers in remote sensing, highlighting their advancements over traditional deep learning methods [27], [28]. Aleissaee et al. in [27] provide a broad survey on Transformer-based approaches, focusing on tasks such as land cover classification, hyperspectral image analysis, and pan-sharpening. While this study demonstrates the potential of Transformers in remote sensing, it primarily examines general image-processing applications and does not specifically address data fusion methodologies for PA.

Similarly, Wang et al. in [28] present a systematic review of Transformers in remote sensing, assessing their performance in segmentation, object detection, and change detection. Although this work evaluates Transformer-based models across multiple remote sensing tasks, it does not extensively explore their role in integrating multi-source agricultural data or improving spatiotemporal fusion techniques in PA.

While these studies contribute valuable insights into Transformers for remote sensing, they do not comprehensively assess their potential for agricultural data fusion. Addressing this gap, this review examines the evolution of data fusion techniques in PA, from foundational methods to state-of-the-art Transformer-based approaches, providing an in-depth analysis of multi-source data integration and its impact on precision agriculture.

This study specifically provides:

1. **A data-driven meta-analysis of remote sensing-based data fusion in pa (1994–2024):** We systematically analyse 30 years of research trends, identifying key turning points in agricultural data fusion and the rise of Transformer-based methods since 2022. Unlike prior studies, which rely on qualitative analysis, we employ a data-driven selection of the top 500 research papers, filtered through chronological and technical relevance criteria.
2. **Comparative evaluation of data fusion techniques, from model-driven to transformer-based methods:** We categorize and compare statistical, machine learning (ML), deep learning (DL), and Transformer-based fusion techniques, assessing their strengths and limitations. Additionally, we specifically evaluate their role in multi-source agricultural data fusion, addressing crop classification, soil analysis, yield prediction, and disease detection.
3. **A structured roadmap for implementing transformer-based data fusion in pa:** We propose a step-by-step framework for applying Transformer-based fusion models in real-world PA applications, covering sensor selection, fusion model design, and validation strategies. This provides an actionable implementation guide, bridging the gap between theoretical advancements and practical deployment in precision farming.

By providing a comprehensive, data-driven analysis, this review offers valuable insights for researchers, policymakers, and agricultural practitioners seeking to advance precision farming through AI-driven remote sensing and data fusion techniques.

## II. RESEARCH METHODOLOGY

Given the extensive volume of publications in this research area, we initially filtered the top 500 papers using citation rankings in Google Scholar, retrieved through the *Publish or Perish* tool. To enhance the rigor and credibility of our selection, we cross-referenced these results with publications indexed in IEEE Xplore and Web of Science, ensuring inclusion of high-impact, peer-reviewed articles from reputable and domain-relevant sources. Google Scholar was particularly valuable as it indexes a wide spectrum of scholarly outputs across platforms such as IEEE, Elsevier, Springer, MDPI, and arXiv. This allowed for a comprehensive and inclusive review, capturing influential research across multiple disciplines and publication types. Our goal was to ensure broad coverage while preserving relevance to the domain of data fusion and remote sensing in precision agriculture, with a particular emphasis on recent Transformer-based advancements.

To systematically extract research trends and key developments, we employed two complementary methodological approaches as follows:

- **Chronological analysis:** This approach examines the evolution of precision agriculture (PA) techniques over time, identifying major methodological shifts, algorithmic innovations, and transitions in remote sensing strategies. A detailed discussion of these trends is presented in Section 0.
- **Data-Driven based statistical analysis:** We conducted a structured analysis of publication trends, identifying breakthroughs, adoption patterns, and thematic shifts within data fusion and AI-based approaches in remote sensing. This approach is elaborated in the following subsection.

This dual-method strategy enabled a comprehensive understanding of how data fusion techniques have progressed in PA, while also highlighting recent innovations that are reshaping the field.

### A DATA-DRIVEN STATISTICAL APPROACH FOR ANALYSING RELATED LITERATURE

The goal of this approach is to extract meaningful patterns, trends and turning points from a large body of literature by analysing multiple dimensions including publication frequency, technique adoption, and journal contributions. The overall workflow is summarized in Figure 1**Error! Reference source not found.** and includes the following steps:

1. **Initial Retrieval**: The top 500 were collected based on citation rankings from Google Scholar using the Publish & Perish tool with the combined keywords such as “data fusion” AND “precision agriculture”.
2. **Visualization and Trend Analysis**: Publication metadata was imported into Power BI to visualize temporal and methodological trends.
3. **Refinement**: We filtered the dataset further by focusing on observable peaks in research activity and aligning them with known technological advancements.
4. **Manual Cleaning**: Duplicates, non-peer-reviewed works, and publications unrelated to agricultural or remote sensing contexts were excluded.

As a result of the data-driven approach based on the top 500 publications within 1994-2024, we extracted several interesting findings. As shown in Figure 2, two significant surges in publication volume were observed: the first in 2019 and the second in 2022. These peaks were used as pivot points to guide our temporal focus as explained in the following.

- The 2019 spike coincides with the broader adoption of Agriculture 4.0, which emerged around 2017 alongside the Industry 4.0 revolution [29]. This era introduced widespread IoT deployments, advanced sensors, and integrated data platforms, which collectively drove increased research in sensor-level data fusion. The uptake of these technologies likely contributed to the observed growth in fusion-related publications during 2019–2020.
- The 2022 peak aligns with the emergence of Transformer-based models in geospatial and remote sensing applications. As shown in Figure 3, Transformer architectures surpassed traditional deep learning methods, such as Long Short-Term Memory (LSTM) networks and Convolutional Neural Networks (CNNs), in terms of adoption for fusion tasks in agricultural contexts beginning in 2022 [28].

- This trend is further confirmed by our domain-specific analysis of IEEE Transactions on Geoscience and Remote Sensing (TGRS), a flagship journal in this field. As illustrated in Table I, out of 1,201 publications related to data fusion in agriculture from 1993 to 2024, 134 papers (11%) used Transformer-based models, and all were published after 2022. This represents a clear methodological shift toward more advanced attention-based architectures.

Based on this evidence, we narrowed our review focus from 2019–2024 to 2022–2024, as this period encapsulates the most technologically significant and rapidly evolving phase in Transformer-based data fusion research. This narrower focus also ensures that our review reflects state-of-the-art and provides forward-looking insights.

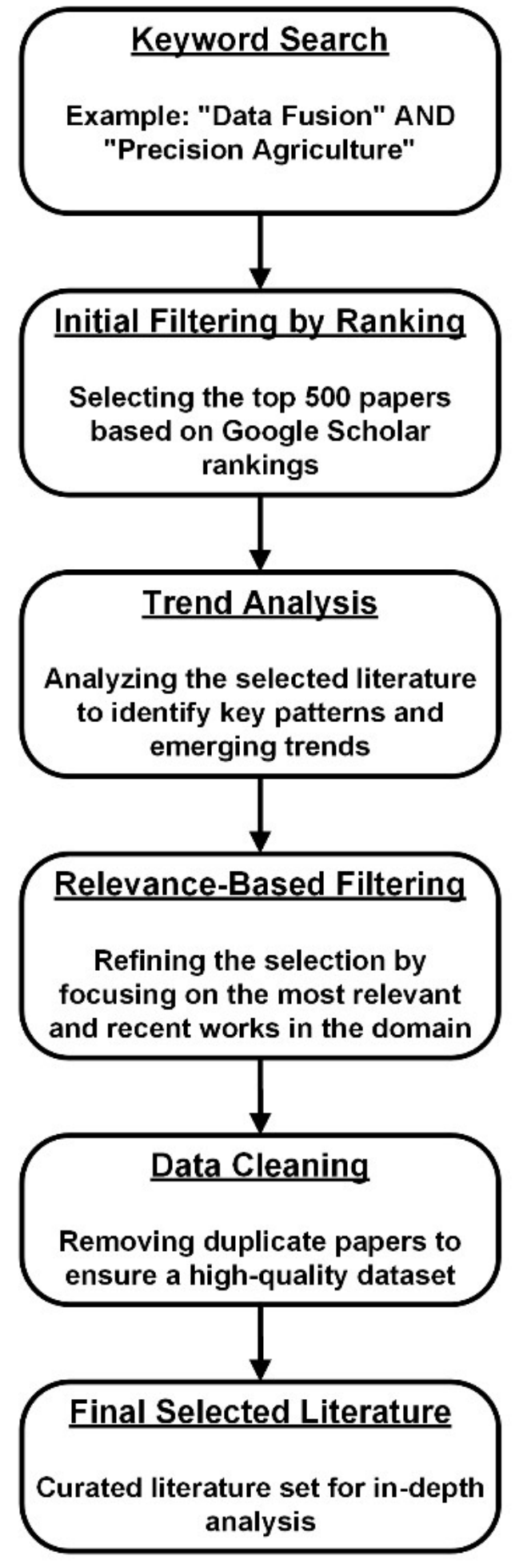

**FIGURE 1.** A data-driven workflow for literature selection, integrating ranking, trend analysis, relevance assessment and refinement stages.

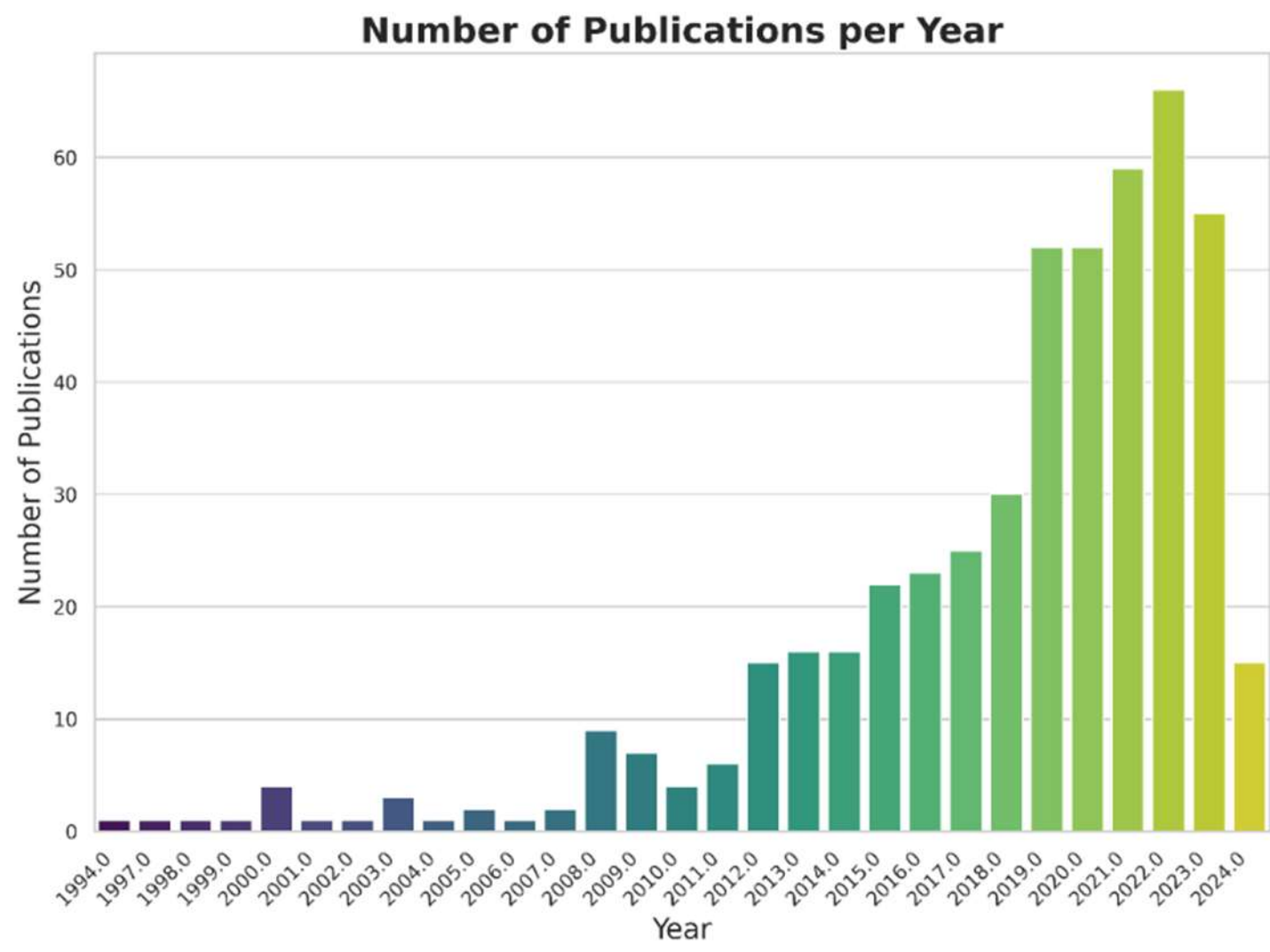

**FIGURE 2.** Annual distribution of publications utilizing data fusion techniques in agriculture (1994–2024).

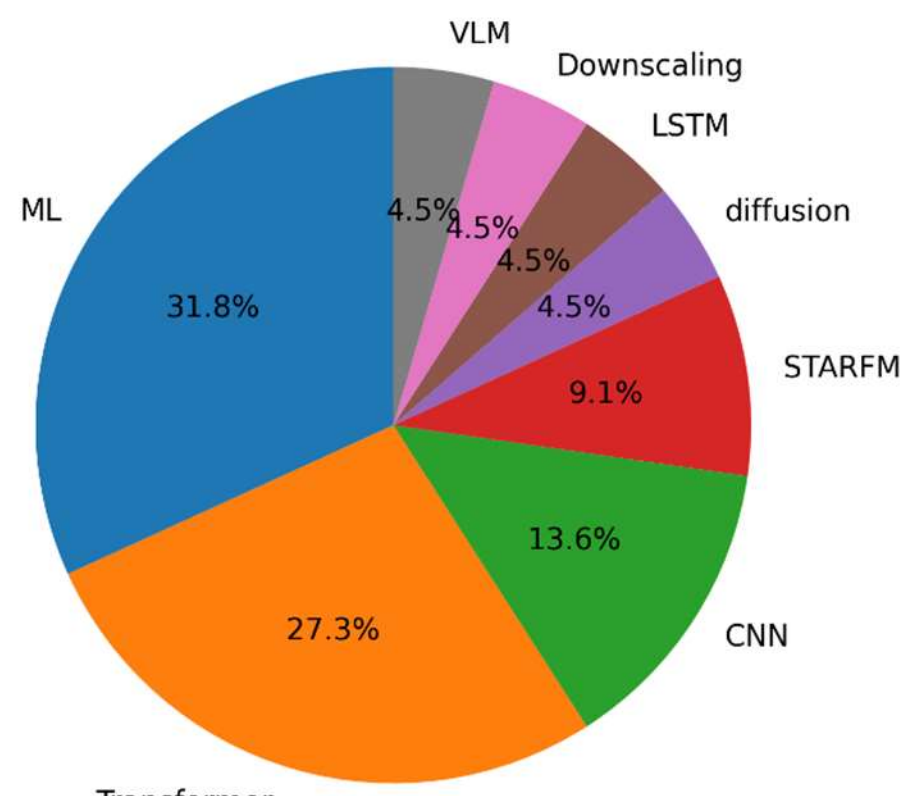

**FIGURE 3.** Comparative analysis of different techniques used in agricultural data fusion.

TABLE I
NUMBER OF DATA FUSION PUBLICATIONS IN IEEE GRS UP TO 2024, CATEGORIZED BY DIFFERENT KEYWORD COMBINATIONS OF "TRANSFORMER" AND "SOIL," ALONG WITH THEIR STARTING YEARS AND CORRESPONDING PERCENTAGES.

| Keywords | No. of Papers | % | Starting Year |
|---|---|---|---|
| Data fusion | 1201 | 100.0 | 1993 |
| Data fusion + Transformer | 134 | 11.2 | 2022 |
| Data fusion + soil | 31 | 2.6 | 1999 |
| Data fusion + Transformer + soil | 4 | 0.3 | 2022 |

*Note:* The percentages are calculated based on the total number of publications (1201).

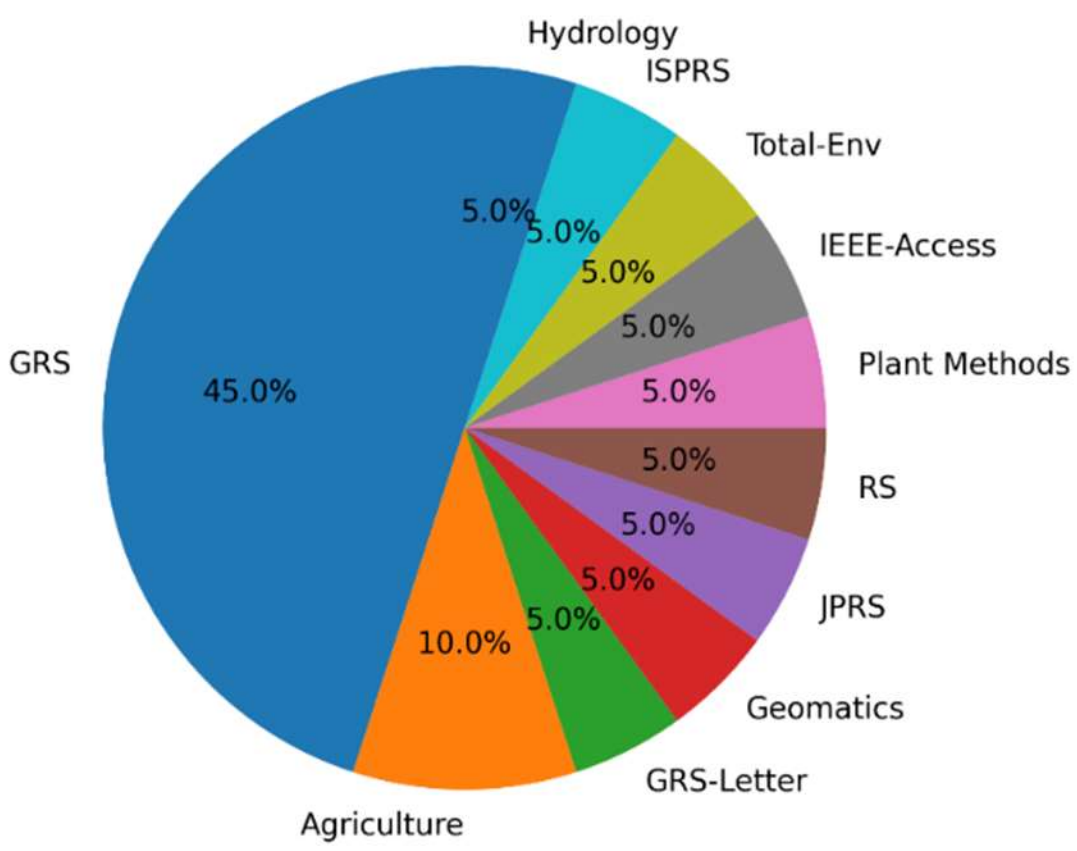

**FIGURE 4. Contribution of different journals to data fusion and remote sensing research.**

## III. PRECISION AGRICULTURE

This section provides an overview of Precision Agriculture (PA), examining its evolution from early research to the latest advancements. PA emerged as a data-driven approach designed to optimize agricultural productivity while minimizing environmental impacts. Early studies established foundational definitions, frameworks, and methodologies focused on managing spatial and temporal variability in agricultural parameters such as crop performance and soil properties [30], [31].

Over time, PA has integrated advanced technologies, including remote sensing, artificial intelligence (AI), the Internet of Things (IoT), and big data analytics, expanding its capabilities in resource management and decision-making [29], [32]. This section explores both the pioneering contributions of early-stage research and the transformative innovations that have reshaped PA into a data-centric and automated field.

### A. PRECISION AGRICULTURE IN EARLY-STAGE LITERATURE

Precision Agriculture (PA) has evolved over the past few decades as a data-driven approach to optimizing agricultural productivity while minimizing environmental impacts. The foundations of PA were laid in the late 20th century, with early research focusing on understanding and managing spatial and temporal variability in crop performance, soil properties, and other key agricultural parameters. Early-stage studies provided essential definitions, frameworks, and methodologies that shaped the development of modern PA techniques. The following section explores these foundational studies, their contributions, and the limitations they faced in the initial stages of PA research.

#### 1) DEFINITIONS

One of the earliest but foundational definitions of PA was provided in [30] at the end of $20^{th}$ century. Based on that, PA is an agricultural approach that uses technologies to evaluate and manage the spatial and temporal variabilities associated with farm productivity to improve crop performance and environmental quality. According to this definition, evaluating and managing temporal and spatial variabilities are key elements of PA as crop performance changes over time and space. Such variability management can be applied to different agricultural parameters including soil fertility (pH, Phosphorus, Potassium, Nitrogen), pest (weed, insect), crop yield and soil moisture (variable rate irrigation, soil-landscape spatial modelling and drainage) [30].

Similarly, as mentioned in [31], the main goal of PA is to improve agricultural production by responding to temporal and spatial variabilities. These variabilities can be classified into 6 groups as follows:

- yield,
- field topography (slope, elevation, etc.),
- soil fertility (nutrients, manure, texture, pH, electric conductivity, density, hydraulic conductivity, soil depth),
- crop (density, height, nutrient stress, water stress, biophysical properties),
- anomalous factors (wind, insect, disease, wind damage, hay damage),
- management (tillage practice, seeding rate, crop hybrid, crop rotation, fertilizer application, pesticide application, irrigation pattern).

#### 2) MANAGEMENT APPROACHES

As mentioned in the previous section, spatiotemporal variable management is one of the key goals of PA. There are 2 different ways to manage these variabilities: the map-based approach (MBA) and the sensor-based approach (SBA) [31]. MBA is easier to implement due to the availability of technologies like remote sensing (RS), GPS and soil and field monitoring. Therefore, most PA experiments are conducted using MBA. However, SBA was less applicable on a larger scale in the early stages due to the high cost of sensors for monitoring soil and field variability. The new trend is to use a hybrid approach that integrates both MBA and SBA.

PA can also help to divide large fields into smaller portions, called management zones (MZ), with homogeneous characteristics [31]. By forming MZs, agricultural inputs can be applied in a site-specific manner. To implement MZs, spatial filtering is required to decrease noise effects from individual measurements.

#### 3) EARLY-STAGE LIMITATIONS

RS had limited applications in PA due to the low resolution of spatial images in the early stages of PA. Early satellite images faced problems such as timeliness, cloud cover, cost, poor quality and lack of sufficient processing power. To address these issues, hyperspectral (HS) imaging emerged as a new approach, with the ability to provide data over a nearly continuous spectrum. Early applications of HS imaging included yield estimation, soil property estimation and crop prediction [31].

### B. PRECISION AGRICULTURE IN RECENT WORKS

As technology continues to advance, Precision Agriculture (PA) has transformed from its early conceptual stages to a highly automated data-driven discipline. Recent studies increasingly emphasize real-time monitoring, intelligent decision-making, and scalable automation driven by AI, IoT, and cloud computing [29], [32], [33]. This subsection synthesizes how recent research redefines PA and identifies key trends and persistent challenges in the field.

#### 1) DEFINITION

The concept of PA evolved significantly with the advent of Agriculture 3.0 in 1992. The introduction of advanced technologies such as AI, the Internet of Things (IoT), and big data in 2017 marked the transition to Agriculture 4.0, establishing a new digital framework. Since then, PA has also been referred to as “smart farming” [29]. Newer definitions incorporate the role of wireless sensor networks, cloud-based analytics, and machine learning for site-specific crop management [32].

Key advances include the use of prescription maps (PMs) that integrate vegetation indices (VIs), soil attributes, and climate data to guide site-specific application of resources such as water, nitrogen, and pesticides. However, many studies still rely heavily on optical indices like Normalized Difference Vegetation Index (NDVI), limiting the richness of decision support tools, especially for soil- or disease-related tasks.

#### 2) SYNTHESIS OF IMPLEMENTATION TRENDS

Recent implementations of PA increasingly leverage remote sensing (RS) in combination with Unmanned Aerial Systems (UAS), in-situ sensors, and edge devices. Satellite-based systems are widely used for scalability, while drones and proximal sensors offer higher spatial resolution but are limited by cost and area coverage. Despite these options, few studies present a clear framework for selecting the appropriate sensing modality based on farm size, crop type, and resource constraints.

Emerging trends include:

- **Multimodal sensing**: Integration of optical, thermal, LiDAR, and hyperspectral data is increasingly applied to overcome the limitations of single-modality systems.
- **On-device intelligence**: New PA systems incorporate edge computing for real-time decision-making directly at the sensor level [34].
- **Data interoperability and cloud connectivity**: Interfacing heterogeneous sensors and legacy systems remains a challenge, especially in developing regions.

Moreover, there is a lack of deployment in less-represented crops like blueberries, tea, or minor legumes, highlighting the need for crop-specific adaptation in PA research
[5].

#### 3) DATA-DRIVEN APPROACHES

The dominance of data-driven models in recent PA research is evident [33]. AI models use multimodal sensor data to estimate critical variables like yield, water stress, and disease presence. A recurring challenge is managing data volume, quality, and missingness, especially when integrating RS with IoT-based field sensors.

Despite these advances, issues such as hardware cost, data heterogeneity, and lack of scalable open-source platforms continue to limit full-scale adoption of PA technologies.

## IV. REMOTE SENSING-BASED DATA FUSION METHODS AND EVOLUTION OF TRANSFORMERS

RS technologies can provide insights on crops and soil properties by capturing data through aircraft, satellites, drones, and ground-based sensors. Images are the most common data collected by RS technologies. RS systems primarily acquire image-based data, which are widely used to generate soil and yield maps. These maps support site-specific management by providing insight into the spatial variability of physical and chemical soil properties, as well as crop productivity [33].

DF plays a critical role in RS by integrating data from multiple heterogeneous sources to improve data quality and decision-making [35]. Initially introduced in the RS literature in 1999 [36], DF encompasses the combination of information across different sensors, spectral channels, and temporal instances. It is typically classified into 3 levels including observation-level, feature-level, and decision-level fusion [37].

Additionally, there are 4 dimensions of information in RS including spatial, temporal, spectral and radiometric. Multitemporal Data Fusion is especially useful for applications with seasonal or rapid changes. Having data with high spatiotemporal resolution is essential for highly dynamic domains like agriculture. AI/ML can predict fine images over time from current coarse images by learning the patterns from the existing coarse-fine image pairs, [38].

A major portion of early-stage works (up to 1998) in RS DF focused on the fusion of RS images for pan-sharpening or the fusion of imagery features for improving classification performances [39]. However, recent research has primarily focused on hyperspectral (HS) images for pan-sharpening purposes [40].

DF plays a major role specially in remote sensing due to the availability of a wide range of heterogeneous sensor types with significantly different temporal, spatial, and radiometric resolutions [41]. However, this has also caused several challenges as follows [42]:

- Data imperfection
- Data inconsistency
- Data conflict: incorrect integration of data from independent processes
- Data alignment/registration and correlation: aligning data collected from various sensors with different frames into a common frame.
- Data type heterogeneity
- Dynamic fusion: ability to fuse data in real or near real time.

DF techniques have been increasingly applied in various domains, including environmental monitoring, urban planning, and particularly in agriculture, where integrating multiple data sources can enhance decision-making and predictive capabilities. Given the complexity of agricultural landscapes and the need for precise monitoring, recent studies have explored innovative ways to leverage DF in remote sensing applications, as explained in the following section.

### A. EVOLUTION OF TRANSFORMERS AND COMPARISON WITH TRADITIONAL DEEP LEARNING APPROACHES

Transformer architectures were originally proposed in natural language processing (NLP) through the seminal model by Vaswani et al. in 2017 [43] , which introduced the self-attention mechanism to effectively model long-range dependencies. This paradigm shift led to the adaptation of Transformers for computer vision tasks, notably through the Vision Transformer (ViT), which demonstrated competitive performance without relying on convolutional structures [44].

In RS and agricultural applications, traditional deep learning models, such as CNNs [45], [46] and recurrent neural networks (RNNs) [47], have long been employed for tasks including image classification, segmentation, and time series analysis. However, Transformer-based architectures offer distinct advantages over these conventional models as follows:

- **Global receptive field:** Unlike CNNs, which rely on localized feature extraction, Transformers attend to all regions of an image or sequence simultaneously, enabling superior spatial awareness [48].
- **Scalability**: Their architecture is inherently parallelizable and well-suited for large-scale datasets, making them ideal for high-dimensional, multi-source RS data [49].
- **Robustness to noise and missing data**: Self-attention mechanisms allow Transformers to focus on relevant features, even when input data is noisy or incomplete [50].
- **Multi-modal fusion capabilities**: Transformers can effectively accommodate heterogeneous data sources such as optical, LiDAR and SAR, supporting comprehensive data integration [51].

### B. APPLICATIONS OF DATA FUSION TECHNIQUES IN AGRICULTURAL REMOTE SENSING

In this section, we analyse the most relevant publications over the last two years, from 2022 to 2024. Table II shows the list of selected publications. We have analysed the publications from different aspects including data type, fusion technique, performance metrics and application.

In terms of fused data, different RS data have been used including SAR, Hyperspectral (HS), Light Detection and Ranging (LiDAR), Infrared (INFRD), visible optical data, such as Red, Green and Blue (RGB), and Multispectral Imaging (MSI). As shown in Figure 6, a combination of SAR and HS has been used more compared to other RS data, with approximately one-quarter of the most recent works.

Regarding the DF techniques, different approaches have been used to combine multimodal RS data in the literature. At a very high-level overview, we can classify the techniques into 2 main groups including model-driven and data-driven approaches. In the model-driven approaches, RS data is combined from multiple sources based on physical models of soil and vegetation indices. Downscaling and STARFM (spatial and temporal adaptive reflectance fusion) techniques used in [52], [53] are examples of model-driven approaches. As shown in Table II, these approaches have received less interest with only 1 relevant work in the recent literature due to their low performances. However, the data-driven approaches including ML and DL have contributed the most due to the recent advances in AI. Support vector machines (SVM) [54], [55], random forest [56], [57], and Gradient Boosting (XGBoost) [58], [59], are examples of ML techniques used in RS DF.

To provide further clarity and transparency in the comparison of fusion strategies, we also classified the reviewed studies in Table II based on three standard fusion levels: (1) Observation-level or pixel-level fusion, where raw RS data from multiple sensors such as SAR and HS are directly combined prior to processing; (2) Feature-level fusion, where extracted representations from each modality are merged before modeling or classification; and (3) Decision-level fusion, where predictions from individual unimodal models are combined through late fusion.

Model-driven methods such as STARFM [63] and downscaling approaches [69] typically perform fusion at the observation level. Most machine learning and deep learning models, including CNNs [47], LSTMs [71], and Transformer-based architectures [56], [60], [59], operate at the feature level, where modality-specific features are integrated prior to prediction. A few works, such as [58], which leverage Vision Language Models (VLMs) with text-image alignment or ensembling, reflect decision-level fusion.

DL techniques include Multilayer Perceptron (MLP) [60], CNN [45], [46], Recurrent Neural Network (RNN) [47], LSTM [60] and Transformers [61], [62], [63], [64], [65], [66], [67]. Compared to other DL techniques, Transformers have received the highest interest in recent years, as illustrated in Figure 3. Therefore, we have allocated a separate category to Transformers to highlight their significance.

As illustrated in Table II, Transformers-based DF techniques demonstrate the highest predictive performance ranging from 92% to 97%, compared to traditional ML and other DL techniques. The median performance metric for Transformer models is 94%, substantially higher than that of ML models at 83% and other DL models at 75%. Furthermore, the interquartile range (IQR), which spans from the first quartile (Q1) to the third quartile (Q3), for Transformer models is as narrow as 3%. By contrast, ML models have an IQR of 11%, and other DL models show an even broader IQR of 28%.

The combination of higher median performance and smaller IQR underscores not only the superior accuracy of Transformer models but also their greater robustness and consistency across different studies. This makes them a particularly promising class of methods for remote sensing-based data fusion in precision agriculture applications. In terms of DF applications in RS, we have classified the works into 5 main categories including land use land cover (LULC) classification [46], [61],

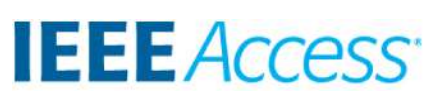

[62], [64], [65], [67], soil prediction [52], [54], [56], [58], [60], [68], crop classification [66], [57], yield prediction [55], [59], and disease classification [45]. Additionally, image interpretation has been recently used as a new application in RS due to the emergence of vision language models (VLMs) [63]. Among the different applications, soil prediction has received higher attention in recent years as shown in Figure 7. Soil prediction addresses a group of applications aimed at estimating various soil parameters including moisture (M) [68], [60], [52] carbon (C) [56], [58], texture [54], and minerals [54] like Nitrogen (N), Phosphorus (P) and Potassium (K) from RS data.

### *A. KEY APPROACHES BASED ON THE LITERATURE*

Transformer-based methods have gained increasing traction in remote sensing and precision agriculture due to their ability to capture long-range dependencies and efficiently integrate multi-modal data sources. Unlike convolutional neural networks (CNNs), which rely on local receptive fields, Transformers leverage self-attention mechanisms to model complex spatial, spectral, and temporal dependencies.

Table III presents recent studies (since 2022) that have applied Transformer-based data fusion (DF) techniques, categorizing them based on model type, application domain, and the nature of input and output data. Recent studies on Transformer-based data fusion can be broadly categorized into 2 primary approaches:

- Approach 1: Transformers as the Primary Data Fusion Model
  - The Transformer directly processes and fuses heterogeneous remote sensing data.
  - Other deep learning models (e.g., CNNs, LSTMs, MLPs) are used for pre-processing or feature extraction.
- Approach 2: Transformers for Feature Extraction and Enhancement
  - Transformers are integrated within the feature extraction pipeline alongside CNNs, attention-based networks, or recurrent architectures (e.g., GRUs, LSTMs).
  - The extracted features from multiple sources are then aggregated using traditional or deep learning-based fusion strategies.

Figure 5 illustrates both approaches, highlighting how Transformers can either serve as the core fusion model or function as part of a hybrid framework. These strategies enable Transformers to effectively capture spatial, temporal, and spectral dependencies, making them particularly advantageous for complex remote sensing applications that require the integration of multi-modal data such as synthetic aperture radar (SAR), hyperspectral imaging (HSI), and LiDAR.

Examples of Approach1 are [69], [70], [71]. Han et al. in [69] proposed AMSDFNet, incorporating deep learning features with an Atrous Spatial Pyramid Pooling (ASPP) module and attention mechanisms for geological interpretation. The network addresses challenges like high interclass similarity and complex spatial distributions by using a hard example mining algorithm. AMSDFNet significantly improves overall pixel accuracy (oPA) and mean intersection over union (mIoU), providing a robust

TABLE II
SUMMARY OF RECENT PUBLICATIONS ON REMOTE SENSING-BASED DATA FUSION TASKS (SINCE 2022). PERFORMANCE METRICS IN THE TABLE REPRESENT ACCURACY (ACC) FOR CLASSIFICATION TASKS AND $R^2$ (COEFFICIENT OF DETERMINATION) FOR REGRESSION (PREDICTION) TASKS.

| Paper | Fused Data | Platform | Technique | Fusion Level | Application | Journal | Year | Performance (%) |
|---|---|---|---|---|---|---|---|---|
| [67] | HS | Aircraft | Transformer | Feature | LULC classification | GRS | 2022 | Acc = 97 |
| [64] | RGB | Satellite | Transformer | Feature | LULC classification | GRS | 2023 | Acc = 97 |
| [61] | HS/LiDAR | Aircraft | Transformer | Feature | LULC classification | GRS | 2024 | Acc = 95 |
| [45] | RGB | Camera | DL | Feature | Disease classification | IEEE-Access | 2022 | Acc = 95 |
| [72] | SAR/HS | Satellite | ML | Decision | Crop classification | Agriculture | 2022 | Acc = 95 |
| [63] | RGB | Satellite/UAV | Transformer | Feature | Image interpretation | GRS | 2024 | Acc = 94 |
| [65] | HS/LiDAR | Satellite/Aircraft | Transformer | Feature | LULC classification | GRS | 2023 | Acc = 94 |
| [62] | INFRD/RGB | Aircraft | Transformer | Feature | LULC classification | GRS-Letter | 2024 | Acc = 92 |
| [66] | SAR/HS | Satellite | Transformer | Feature | Crop classification | RS | 2023 | Acc = 92 |
| [46] | HS/LiDAR/SAR | Aircraft | DL | Feature | LULC classification | GRS | 2022 | Acc = 88 |
| [57] | RGB/INFRD/HS | UAV | ML | Feature | Crop classification | Plant Methods | 2022 | Acc = 78 |
| [53] | HS/MS | Satellite | Model-driven | Pixel | Downscaling | GRS | 2024 | Not applicable |
| [73] | SAR | Satellite | Transformer | Feature | Soil element prediction | IEEE-Access | 2024 | $R^2$ = 99 |
| [54] | HS | Satellite | ML | Pixel | Soil element prediction | Geomatics | 2024 | $R^2$ = 91 |
| [68] | SAR/HS | Satellite | ML | Feature (Scheme 1)<br>Decision (Scheme 2) | Soil element prediction | GRS | 2024 | $R^2$ = 89 |
| [58] | SAR/HS | Satellite | ML | Feature | Soil element prediction | Total-Env | 2022 | $R^2$ = 87 |
| [59] | HS/LiDAR | UAV | ML | Feature | Yield prediction | ISPRS | 2022 | $R^2$ = 79 |
| [52] | INFRD/RGB | Satellite | Model-driven | Decision | Soil element prediction | GRS | 2024 | $R^2$ = 74 |
| [55] | RGB/INFRD/HS | UAV | ML | Feature | Yield prediction | Agriculture | 2023 | $R^2$ = 60 |
| [60] | SAR/HS | Satellite | DL | Decision | Soil element prediction | Hydrology | 2022 | $R^2$ = 60 |

TABLE III
SUMMARY OF RECENT PUBLICATIONS USING TRANSFORMERS AS A KEY COMPONENT IN DATA FUSION ARCHITECTURES (SINCE 2022).
(ISD: INTERPRETATION SYMBOL DISTANCE, ASTER: ADVANCED SPACEBORNE THERMAL EMISSION AND REFLECTANCE RADIOMETER)

| Reference | Year | Model | Application Area | Data Inputs | Results/Outputs |
|---|---|---|---|---|---|
| [74] | 2024 | ViT + GAN | Soil element classification | Radar | Different soil types (sand, silt, clay) |
| [75] | 2024 | Ensemble dense inception Transformer | Seismic horizons classification | Radar | Seismic horizons (7 layers) |
| [69] | 2022 | CNN + Transformer | LULC classification | Multispectral | LULC categories (13 classes) |
| [70] | 2024 | Temporal Fusion Transformer | Forecasting soil water potential | Soil sensor data | Soil water potential |
| [71] | 2024 | Multimodal Transformer | LULC classification | RGB, SAR, DSM, MS, LiDAR | LULC categories (37 classes) |
| [76] | 2024 | Gated Recurrent Unit (GRU) and Transformer | Soil moisture forecasting | Soil and meteorological data (temperature, humidity, precipitation, wind speed) | Soil moisture content at 6 different depths (10cm – 60cm) |

*ISD: The minimum Euclidean distance from a given point to the edge of the water body is calculated as the ISD for that point.

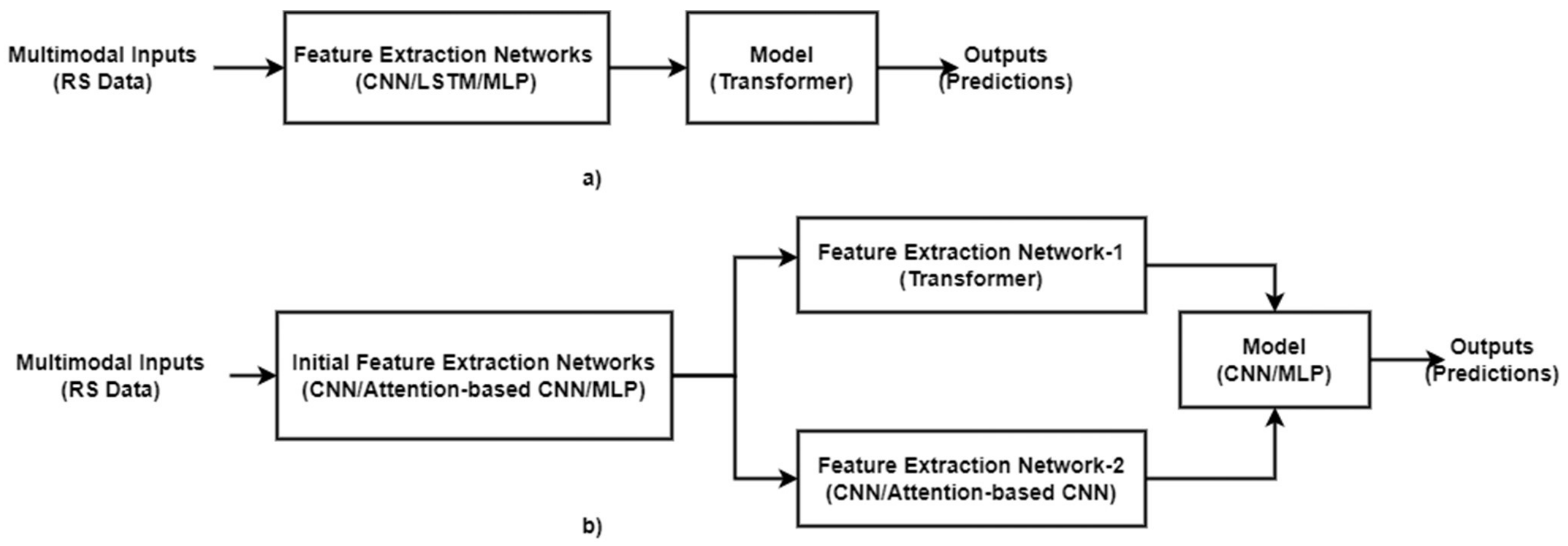

**FIGURE 5. Overview of Transformer-based data fusion architectures: a) Approach 1, where Transformers are used as the primary model; b) Approach 2, where Transformers are employed partially for feature extraction.**

## V. TRANSFORMER-BASED DATA FUSION STRATEGIES

As another example of Approach1, a multimodal Transformer network has been used in [71] to address the challenge of multimodal data fusion with missing modalities. The model incorporates learned fusion tokens, modality attention, and masked self-attention mechanisms, supporting both supervised and self-supervised pretraining. The framework allows integration of incomplete multimodal inputs during training and inference, improving robustness and performance in tasks like building instance/semantic segmentation and land-cover mapping.

Examples of Approach1 are [74], [75] and [76]. EDIFormer was introduced in [75], which integrates Dense Extreme Inception Network (DexiNet) with a Transformer module for automating seismic horizon picking. By leveraging multiple seismic attributes, the model enhances the accuracy and reliability of horizon predictions. The ensemble learning approach combined various models trained on different attributes, outperforming traditional CNN-based methods in providing smooth and accurate horizon interpretations.

In RS applications, there are situations where acquiring extensive data is challenging. In such situations, employing a method to generate data that closely resembles real-world scenarios can be vital. That is the topic in [74], which proposed the FM-GAN, integrating a conditional generative adversAerial network (CGAN) with a vision Transformer (ViT) for generating realistic B-scan images in ground-penetrating radar (GPR) applications. The model's adaptive spatial polarization generator and double SimAM feature fusion attention module (DSFFA) significantly enhance image generation. The FM-GAN outperforms traditional methods in computational efficiency and fidelity, providing a promising tool for geological exploration and subsurface target detection.

The key strength of Transformers is their ability to capture long-term dependencies. By combining with other methods, we can improve Transformer's ability in capturing short-term dependencies as well. That is the idea behind [76]. The authors introduced an innovative hybrid model that integrates Gated Recurrent Unit (GRU) and Transformer technologies to improve the accuracy of soil moisture content forecasts. Using data from eight monitoring stations in Hebei Province, China, from 2011 to 2018, the study assesses the model's performance against various input variables and forecast durations. The results show that the GRU–Transformer model excels in short-term projections (1- to 2-day latency) with a mean square error

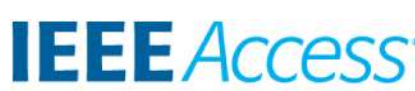

(MSE) of 5.22% for a 1-day forecast, reducing to 2.71%, and a mean coefficient of determination (R2) of 89.92%. The model demonstrates versatility in managing different soil depths, performing exceptionally well at greater depths. It also shows resilience to varied parameter configurations, maintaining high accuracy with a reduced set of soil moisture content parameters. Compared to other prevalent models, the GRU–Transformer framework significantly enhances prediction precision for soil moisture content, providing a robust decision-support tool for agricultural irrigation planning and contributing to water resource conservation and optimization in agriculture.

### B. MATHEMATICAL FORMULATION AND KEY ARCHITECTURES

#### 1) MATHEMATICAL FORMULATION

Transformers have emerged as a unifying framework for learning structured dependencies in high-dimensional, multi-modal data, making them particularly suitable for remote sensing-based data fusion. In this context, we consider a theoretical framework grounded in multi-modal representation learning, where the goal is to integrate heterogeneous data sources including spatial, spectral, temporal, and auxiliary static features into a shared latent space that preserves task-relevant information while discarding modality-specific noise.

The architecture of the Transformer model illustrated in Figure 9 is based on a stack of encoder and decoder layers, where each encoder-decoder pair facilitates cross-modal interactions via attention mechanisms. Let the input $X$ be a sequence of modality-aware embeddings:

$$\boldsymbol{X} = [\boldsymbol{x_1}, \boldsymbol{x_2}, \dots, \boldsymbol{x_n}] \in \mathbb{R}^{\boldsymbol{n \times d}} \quad (1)$$

where $x_i$ represents the fused feature embedding from multiple sources at timestep $i$, and $d$ is the embedding dimension.

Each Transformer block consists of the following components [43]:

1. **Multi-Head Self-Attention (MHSA):** Encodes pairwise relationships between features and positions in the sequence:

$$\boldsymbol{MHSA(X) = Concat(head_1, head_2, \dots, head_h)} \quad (2)$$

where $h$ represents the number of attention heads, and each head is computed as:

$$\boldsymbol{head_i = Attention(Q, K, V) = softmax\left(\frac{QK^T}{\sqrt{d_k}}\right)V} \quad (3)$$

with:

$$\boldsymbol{Q = XW_Q, K = XW_K, V = XW_V} \quad (4)$$

where:

- $Q, K, V$ are the query, key, and value matrices obtained by applying learned weight matrices $W_Q, W_K, W_V$.
- $d_k$ is the scaling factor to stabilize gradients.
- The Softmax function normalizes attention weights.

2. **Residual Connections and Layer Normalization:** Applied after each sub-layer to stabilize training.
3. **Positional Encoding:** Since the Transformer lacks inherent recurrence or convolution, positional information is injected as:

$$PE_{(pos,2i)} = sin\left(\frac{pos}{10000^{2i/d}}\right)$$
$$PE_{(pos,2i+1)} = cos\left(\frac{pos}{10000^{2i/d}}\right)$$

Figure 10 illustrates the attention mechanisms in detail.

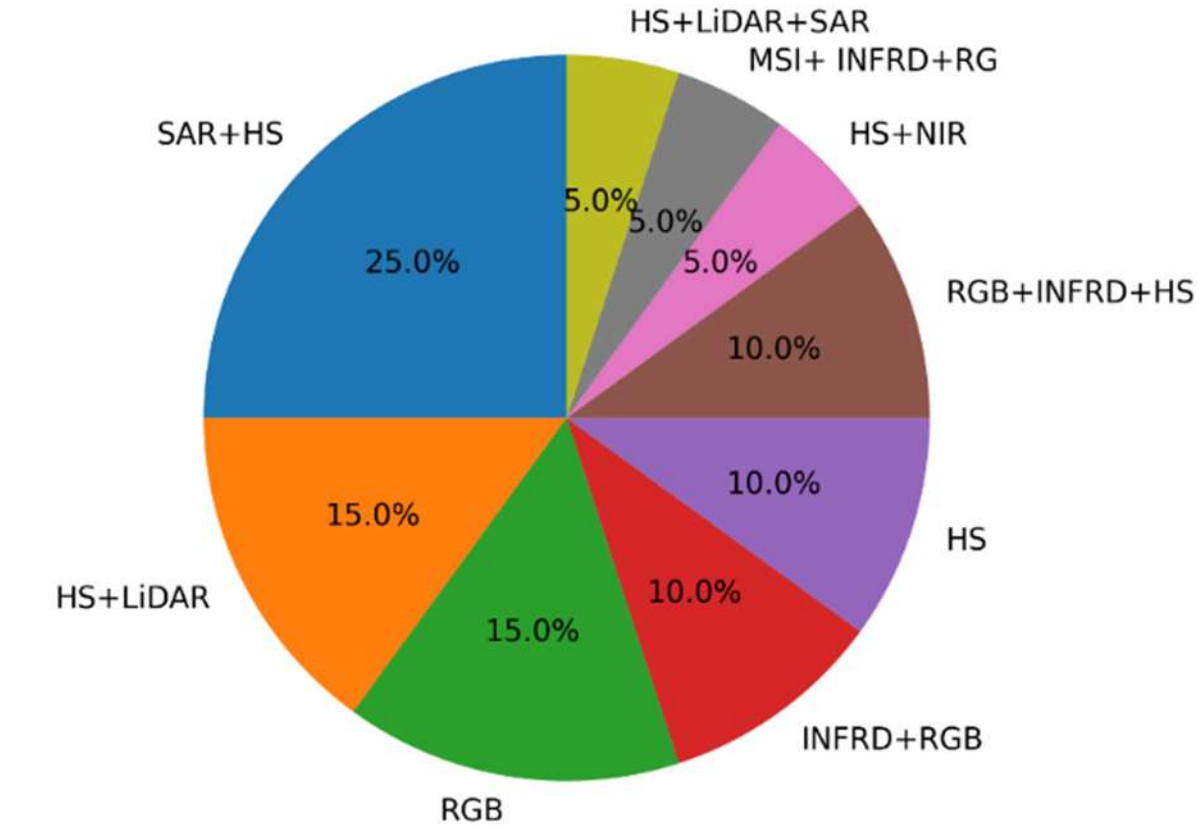

**FIGURE 6. Distribution of remote sensing data types used in data fusion across the selected publications.**

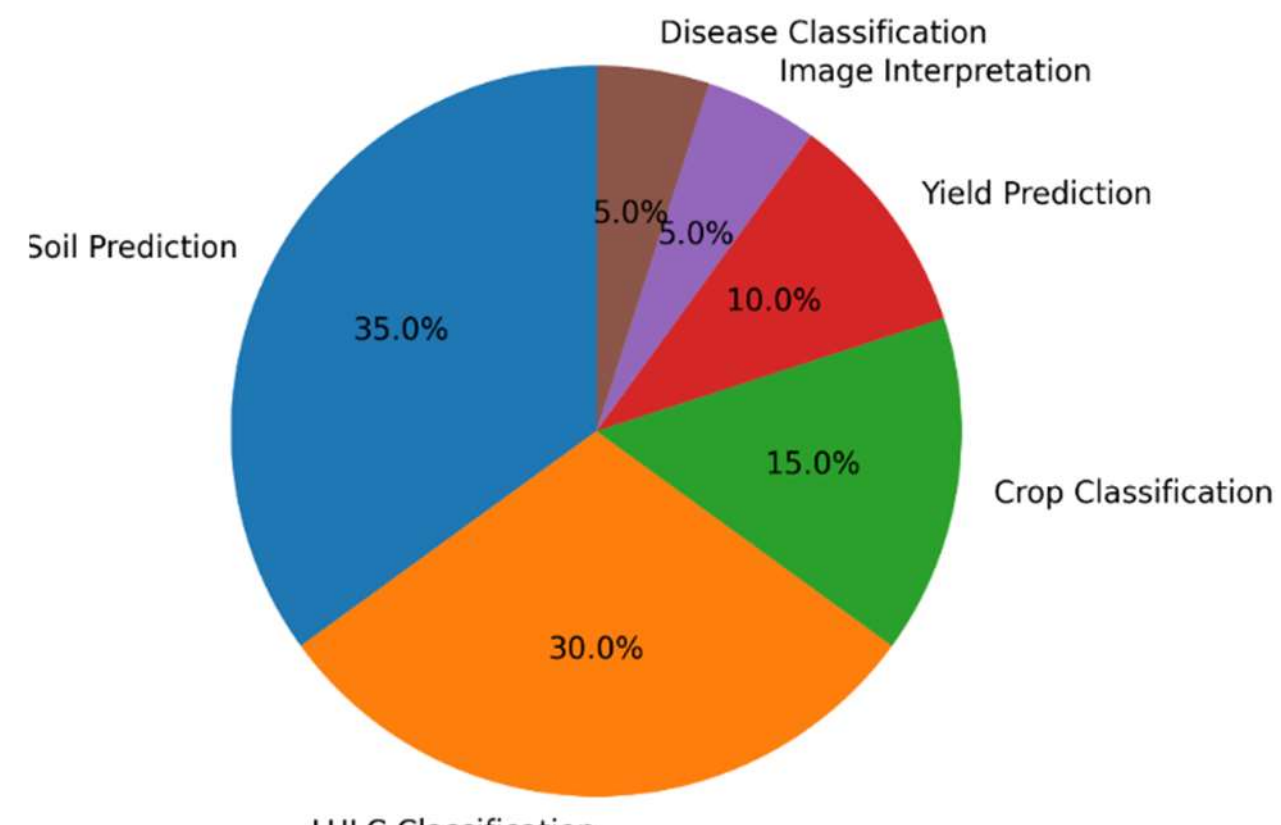

**FIGURE 7. Distribution of data fusion applications across the selected literature.**

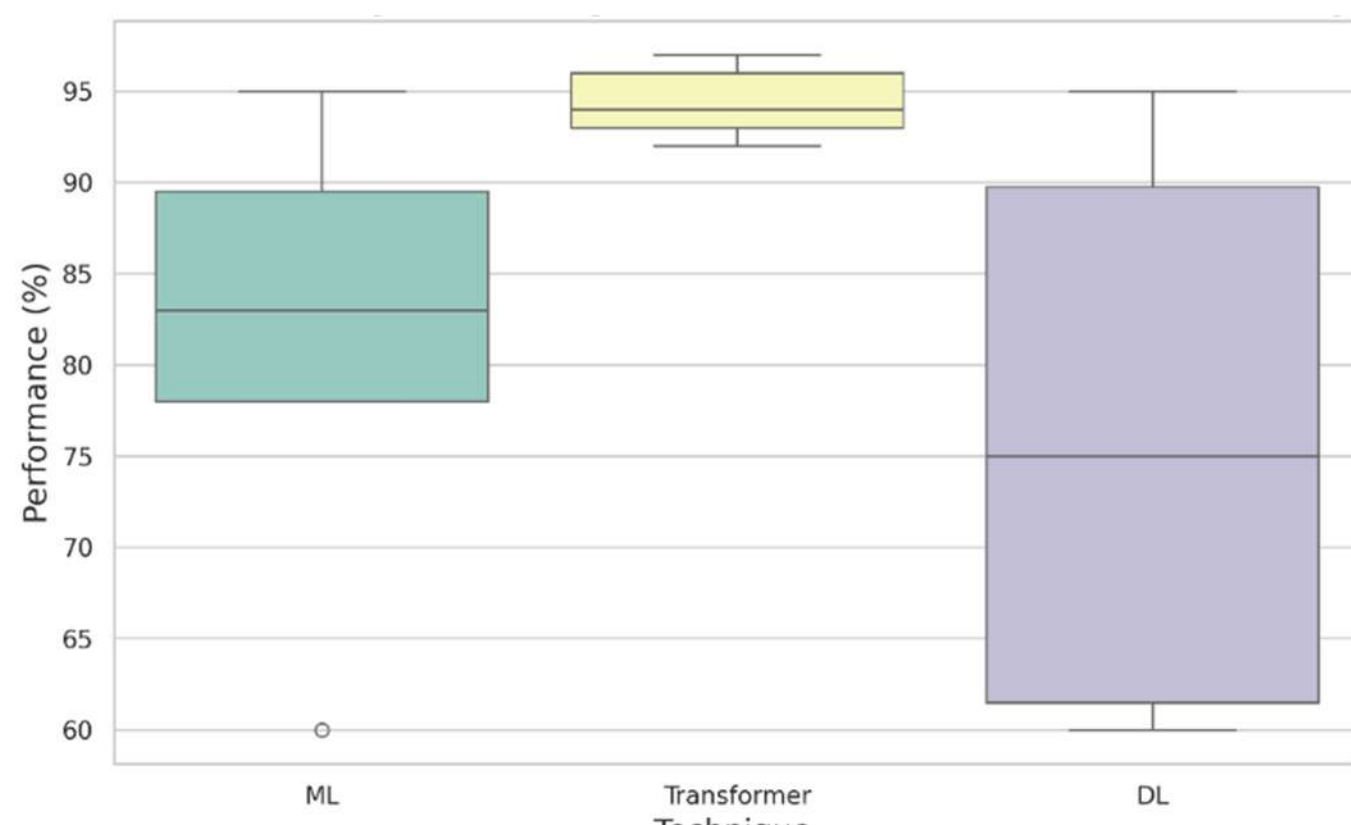

**FIGURE 8. Performance of the data fusion process using various AI techniques, including machine learning (ML), deep learning (DL), and Transformers. The performance metrics is accuracy for classification and $R^2$ for regression tasks.**

#### 2) BASIC MODELS IN REMOTE SENSING

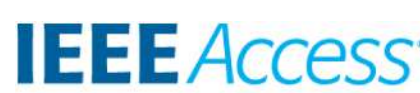

Several Transformer architectures have been adapted for remote sensing applications, particularly for precision agriculture. Some notable examples include:

a) Vision Transformer (ViT) for Multi-Resolution Data Fusion

The Vision Transformer (ViT) is a Transformer-based model that directly processes sequences of image patches for image classification [44]. It closely adheres to the original Transformer architecture. The overall framework of ViT is shown in Figure 11.

ViT has been widely used for integrating multispectral (MSI) and hyperspectral (HSI) data, where its patch-based processing allows effective learning of spatial features across different resolutions [77].

Unlike CNNs, which rely on kernel-based feature extraction, ViTs encode global dependencies, making them highly effective in handling spectral distortions in satellite imagery [77].

b) Temporal Fusion Transformer (TFT) for Time-Series Forecasting

TFT models long-term dependencies in sequential data, making them highly effective in soil moisture prediction and yield forecasting [70].

Compared to LSTMs and traditional autoregressive models, TFTs improve generalization by leveraging variable selection layers that dynamically prioritize the most relevant input features **[78]**. Figure 12 Illustrates the architecture of a TFT network. The TFT takes as input static metadata, time-varying historical data, and time-varying future inputs known a priori. To enhance predictive performance, a Variable Selection mechanism identifies the most relevant features dynamically. Gated Residual Network (GRN) blocks facilitate efficient information flow through skip connections and gating mechanisms. Temporal dependencies are modelled using LSTMs for local sequence processing, while multi-head attention integrates information across different time steps.

### 3) HYBRID MODELS: ENHANCING TRANSFORMER PERFORMANCE

While Transformer models are powerful in capturing long-range dependencies and complex spatial-spectral interactions, they often face limitations in modelling local structures or short-term temporal dynamics, especially in fine-grained agricultural datasets. To overcome these constraints, researchers have proposed several hybrid architectures that integrate Transformers with Convolutional Neural Networks (CNNs), Recurrent Neural Networks (RNNs), or Graph Neural Networks (GNNs). These models aim to leverage the complementary strengths of different architectures, both in terms of predictive performance and deployment efficiency.

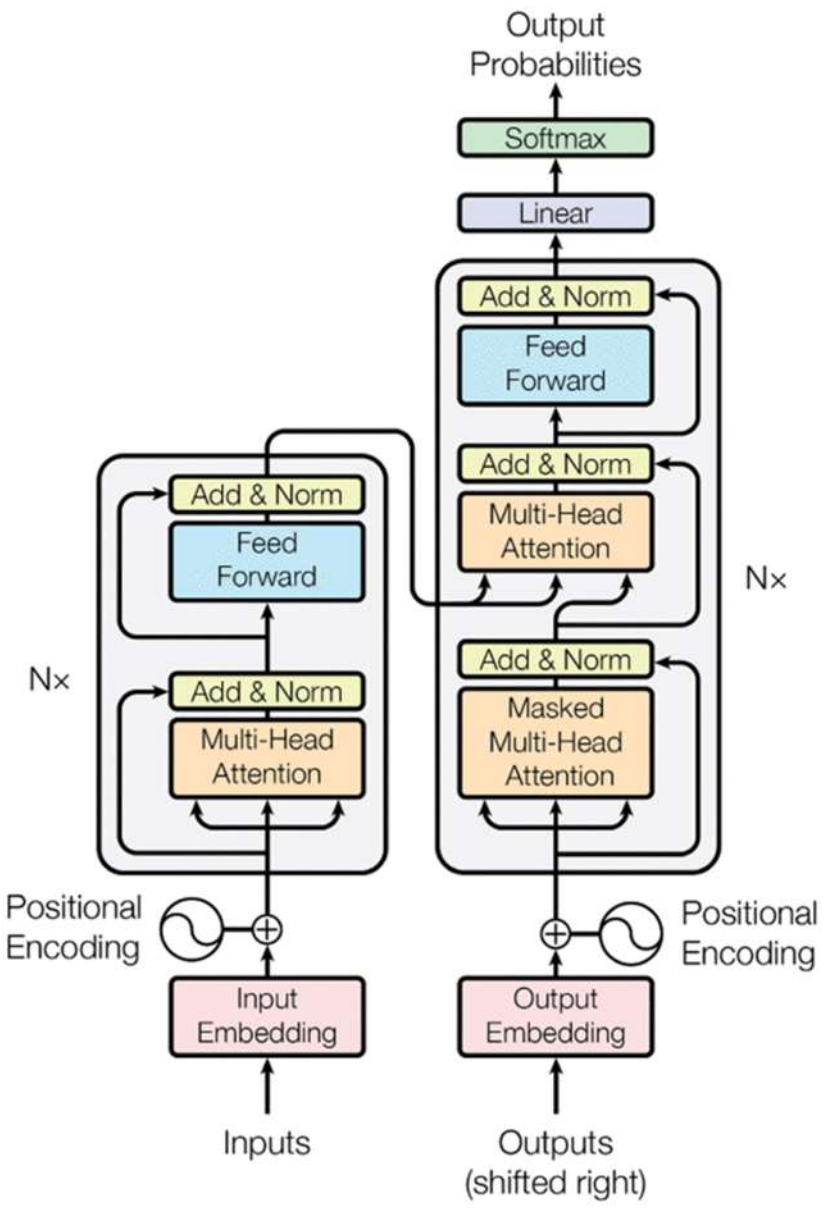

**FIGURE 9. The architecture of an original Transformer model [43].**

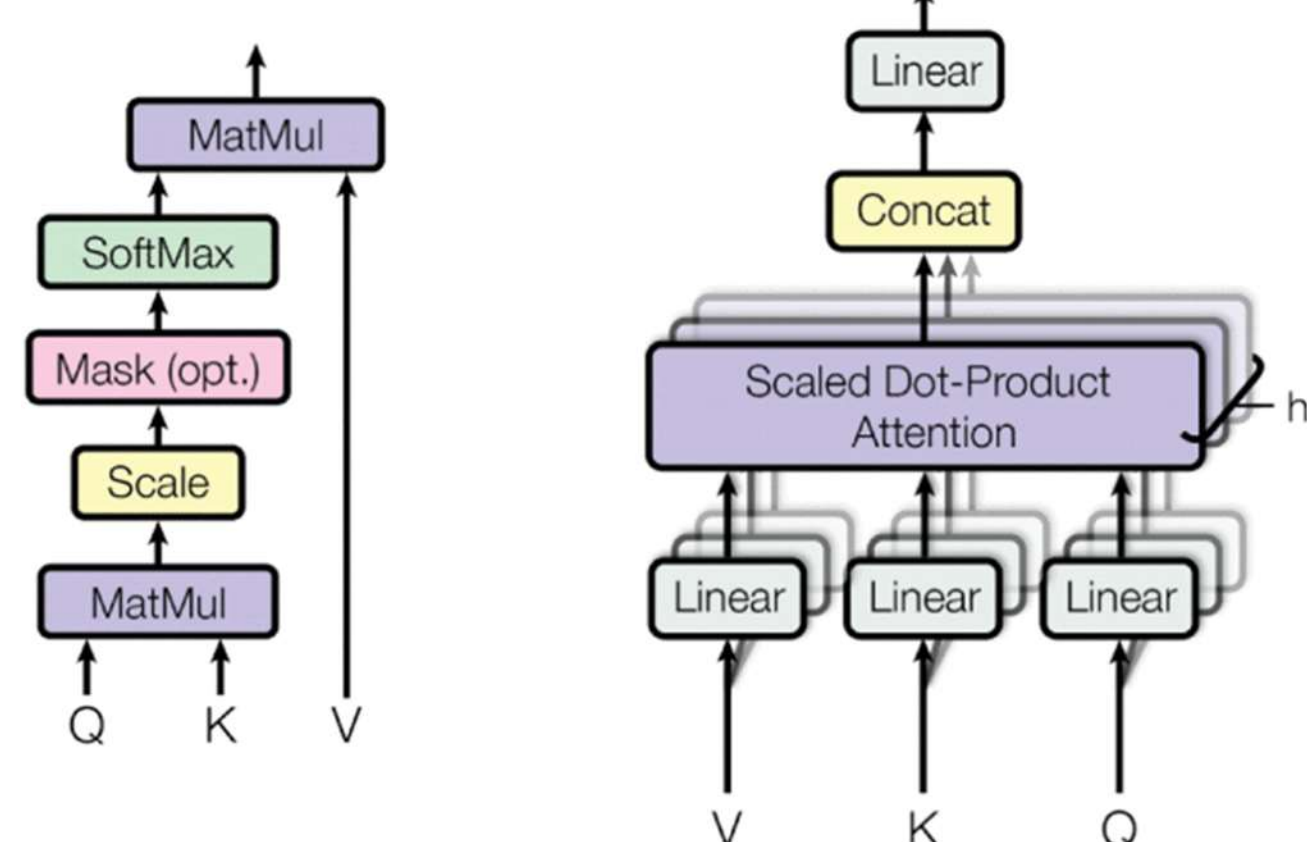

**FIGURE 10. Attention mechanisms in Transformers: The left diagram illustrates the self-attention mechanism, while the right diagram depicts the multi-head attention mechanism [43].**

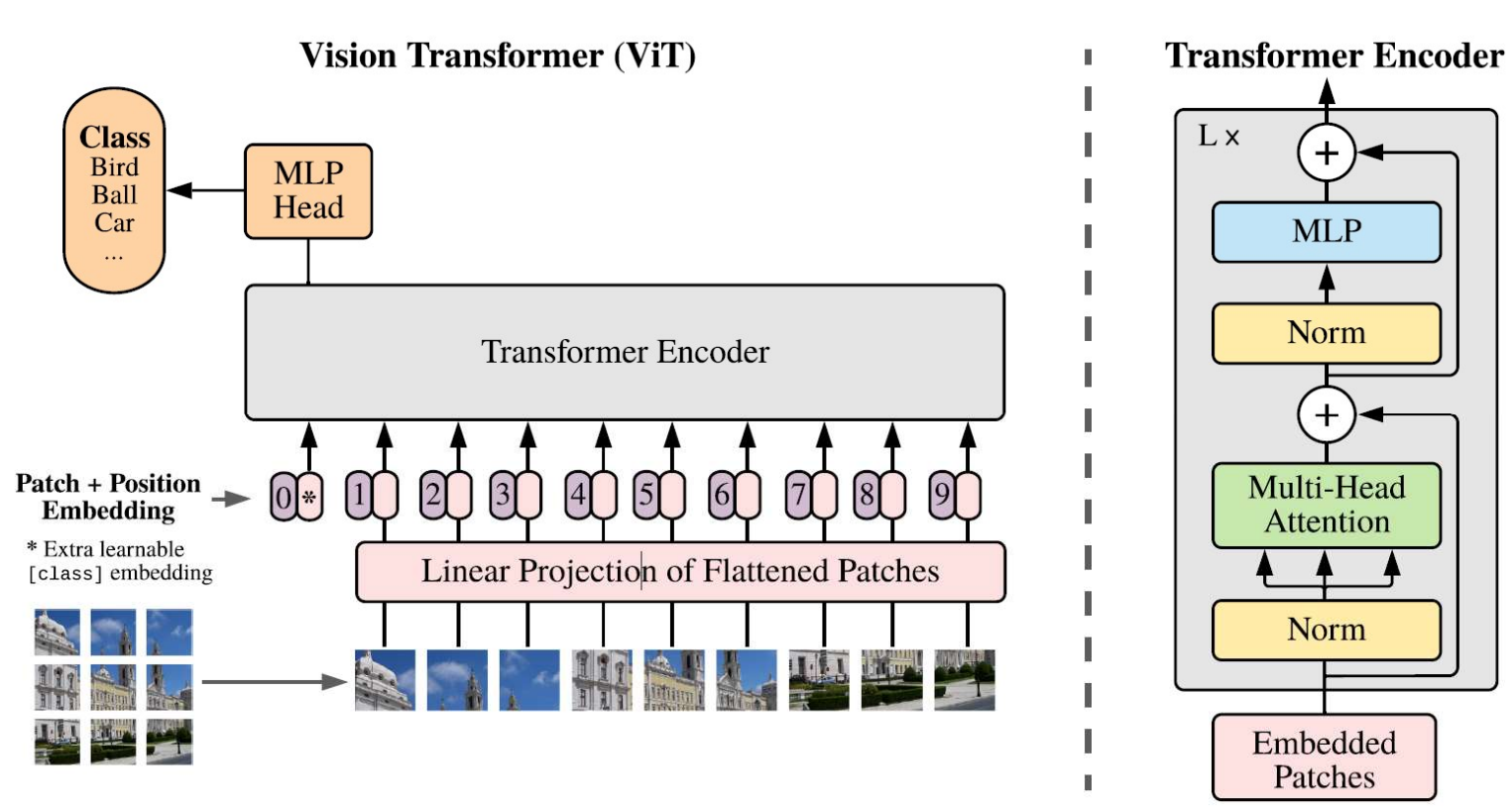

**FIGURE 11. The ViT architecture [44].**

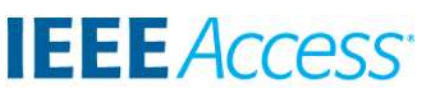

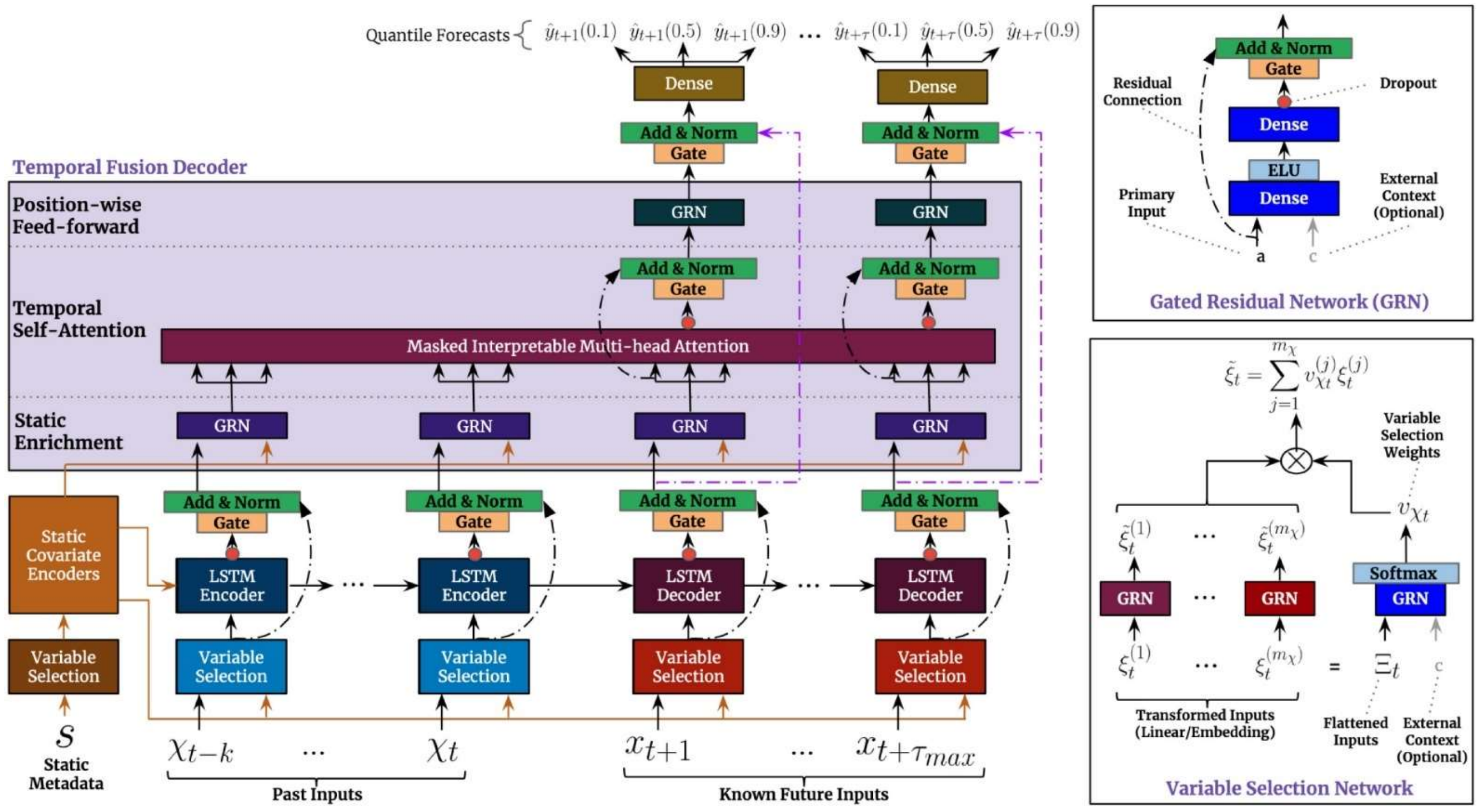

**FIGURE 12.** The architecture of a Temporal Fusion Transformer [78].

TABLE IV
COMPARISON BETWEEN TRANSFORMER-BASED AND TRADITIONAL METHODS IN AGRICULTURAL DATA FUSION

| Aspect | Traditional Methods (e.g., RF, CNN, LSTM) | Transformer-Based Methods |
|---|---|---|
| Accuracy and prediction power | Moderate to high; CNNs and LSTMs effective for spatial/temporal patterns | Very high, especially for multimodal, long-sequence data fusion |
| Computational requirements | Lower to moderate; feasible on standard hardware | High due to quadratic complexity; requires GPU/TPU or cloud-based resources |
| Model interpretability | High for RF; moderate for LSTM/CNN | Low; requires attention visualization or explainable AI techniques |
| Training data requirements | Modest; traditional models often effective with small to medium datasets | Large datasets typically needed; self-supervised learning (SSL) can mitigate this |
| Scalability to high-resolution data | Tree-based models like RF offer lightweight scalability, but deep learning models like CNNs and LSTMs face limitations in handling high-resolution data. | Scalable with advanced variants like Swin Transformer and Linformer |
| Adaptability to new tasks | Moderate; retraining often required | High via transfer learning and fine-tuning on related tasks |
| Deployment in low-resource settings | More feasible; lightweight versions can run on edge devices | Challenging; requires model compression or distillation for deployment |
| Robustness to missing/noisy data | RF is robust; LSTM/CNN performance may degrade | Sensitive to quality but can be improved with fusion of multimodal inputs |
| Fusion capability (multi-source) | Limited; requires feature engineering or stacking | Strong; attention mechanism enables dynamic weighting of heterogeneous sources |

a) CNN-Transformer for Local-Global Feature Integration

CNNs are particularly effective at capturing localized spatial features such as texture, edges, and patterns from high-resolution multispectral or hyperspectral imagery. In CNN-Transformer hybrids, CNN layers typically act as a front-end feature extractor, feeding refined spatial embeddings into Transformer encoders for global context modelling.

For instance, in crop classification and weed detection tasks, CNN blocks (e.g., ResNet or MobileNet) are used to encode high-frequency textures, which are then passed to Transformer layers to model inter-class spectral relationships [81]. This two-stage design has demonstrated improved accuracy and data efficiency compared to either model in isolation. Recent adaptations such as Swin-Transformer also integrate hierarchical CNN-like feature extraction within the attention architecture itself, facilitating multiscale representation learning suitable for field-level analysis.

b) RNN-Transformer for Temporal Forecasting

In sequential prediction tasks like soil moisture estimation,

evapotranspiration modelling, or disease forecasting, short-term temporal dependencies are crucial. RNNs such as Gated Recurrent Units (GRU) and LSTM are well-suited for capturing local trends and recency effects, which can complement the Transformer's global self-attention mechanism [76]. For example, Zheng et al. [76] proposed a a GRU-Transformer architecture for soil moisture estimation that outperformed Transformer-only, LSTM-only, and XGBoost models by 2–6% in $R^2$. Similarly, integrating LSTM layers with Transformers has been shown to enhance prediction stability over longer lag periods and under data sparsity conditions [82]. These hybrid models are particularly advantageous when high-frequency environmental data such as daily or sub-daily weather is fused with lower-resolution satellite imagery.

c) Deployment Considerations: Scalability and Interoperability

Despite their predictive power, Transformer-based models (especially hybrids) can be computationally intensive. For deployment in real-world agricultural systems, scalability and interoperability must be addressed.

**Scalability**: Hybrid models can be adapted for lightweight deployment through architecture simplification (e.g., shallow CNN-Transformer blocks), model compression techniques (e.g., pruning, quantization), or use of edge-efficient variants like TinyViT or Mobile-Transformer. These enable deployment on edge devices such as drones, proximal sensors, or smart tractors with limited compute resources.

**Interoperability**: Integration with existing farm systems requires that models can ingest data from heterogeneous sources (e.g., IoT sensors, yield monitors, remote sensing platforms). This can be achieved through modular preprocessing pipelines and adherence to agricultural data standards such as ODK (Open Data Kit) or OADA (Open Ag Data Alliance) protocols. Additionally, containerized deployment (e.g., via Docker) and REST-based model serving allow seamless integration into farm management software. Computational Challenges

While Transformer-based fusion methods offer state-of-the-art performance in many precision agriculture applications, several practical and computational challenges remain—particularly in resource-constrained settings common to developing regions and smallholder farms.

### C. COMPUTATIONAL COMPLEXITY

Traditional models such as Random Forests, SVMs, and CNNs are often more efficient and feasible for deployment on local servers or edge devices. Transformers require quadratic memory with respect to input sequence length, posing significant challenges when working with high-resolution satellite imagery or dense temporal sequences [83]. Lightweight variants such as Linformer [84] and Swin Transformer offer hierarchical or low-rank attention mechanisms that partially alleviate these costs. Nevertheless, the memory and compute demands still exceed those of conventional models, requiring high-end GPUs or cloud infrastructure, which may not be accessible in every agricultural region.

- **Cost-Benefit Trade-offs:** While Transformer-based models can significantly improve yield prediction, soil moisture estimation, and disease detection through better spatiotemporal representation, these gains must be weighed against energy, hardware, and latency costs. For instance, field-deployable solutions may prioritize explainable models with fast inference such as decision trees or hybrid CNN-LSTM networks, even at the expense of marginal performance gains.
- **Data Scarcity:** Transformer models typically require large-scale labelled datasets, which are often unavailable in agricultural domains. Self-supervised learning (SSL) and contrastive learning offer promising alternatives for pretraining models using unlabelled multimodal data [87], but their adoption remains limited due to implementation complexity. In contrast, traditional models and shallow learning techniques can perform reasonably well on small datasets with proper feature engineering.
- **Explainability and Interpretability:** While Transformers provide superior accuracy, their decision-making processes are less interpretable than traditional models. Attention maps and saliency-based methods can improve model transparency for practical deployment [85].

To further clarify the strengths and limitations of Transformer-based methods in comparison to conventional approaches, Table IV summarizes key differences across multiple dimensions relevant to agricultural applications, including resource constraints, scalability, and deployment feasibility.

By effectively capturing spatial, spectral, and temporal dependencies, Transformers enhance the integration of heterogeneous data sources, improving prediction accuracy and robustness. While Transformers serve as powerful standalone fusion models, their performance is further enhanced when combined with complementary approaches such as CNNs, LSTMs, and generative models. However, the successful implementation of these methods relies not only on the choice of model but also on the underlying data sources and integration strategies. This brings us to the broader data fusion process, which encompasses the selection of data sources, fusion methodologies, and their practical applications. These factors are embedded in our roadmap recommendations (Section VIII), which differentiate between research-focused and deployment-oriented pathways, offering actionable guidance based on computational feasibility, interpretability, and model accuracy.

## VI. DATA FUSION PROCESS

Data fusion (DF) is a fundamental technique in remote sensing and agricultural applications, enabling the integration of heterogeneous data from multiple sources to enhance decision-making and analytical accuracy [86]. By combining complementary datasets—such as optical imagery, synthetic aperture radar (SAR), and LiDAR—DF mitigates the limitations of individual sensors, leading to improved spatial, temporal, and spectral information extraction [87]. The growing availability of multi-source remote sensing data and advancements in machine learning have further expanded the potential of DF, making it an essential approach for various applications, including precision agriculture, environmental monitoring, and resource management [88].

The data fusion process consists of three fundamental components including data sources, integration techniques, and target applications [42]. The application of data fusion in agriculture was described in Section-IV.A, emphasizing its role in enhancing crop monitoring, yield prediction, and soil analysis. This section provides an overview of the key data sources and integration techniques, highlighting their significance in DF. Building upon this foundation, we present a structured roadmap that systematically combines these elements, offering a strategic framework for effective DF implementation across diverse domains.

### A. DATA SOURCES AND GROUND TRUTH SELECTION

The first step in data fusion (DF) is identifying the data sources necessary for integration. As illustrated in Figure 6, remote sensing (RS) data can be categorized into various modalities, each offering unique advantages for environmental monitoring and analysis. These include optical, infrared, radar, LiDAR, multispectral (MS), and hyperspectral (HS) imagery [89]. Each modality provides different levels of spatial, spectral, and temporal resolution, and their effectiveness is determined by the specific requirements of the research and the type of environmental features being studied.

However, the utility of these modalities heavily relies on the availability of accurate and representative ground truth data, which is crucial for the calibration and validation of remote sensing models [90]. Ground truth data serves as a reference that enables the assessment of the accuracy of RS-derived products. This data can include field-measured parameters such as soil properties, crop biomass, and observations from weather stations [91], [92], [93]. However, challenges exist in acquiring high-quality ground truth data due to factors such as limited access to field sites, variability in data quality, and discrepancies in spatial or temporal resolution between remote sensing and ground truth measurements [94].

Numerous studies emphasize the need for robust ground truth datasets to calibrate and validate remote sensing models. For example, soil properties, such as moisture content and texture, are critical for improving the accuracy of vegetation indices derived from optical and infrared imagery. Similarly, weather station data provides essential context for interpreting temporal changes in vegetation or land surface characteristics observed via radar or LiDAR. The integration of such diverse data sources enhances the precision of remote sensing models and supports their application in real-world scenarios, from precision agriculture to environmental monitoring.

### B. DATA FUSION TECHNIQUES

Once the data sources are selected, an appropriate data fusion technique must be chosen to integrate and analyse the data effectively. Various data integration methods are available, with machine learning (ML) and deep learning (DL) techniques being among the most prominent [95]. Traditional fiML methods, such as Support Vector Machines (SVMs), Random Forest (RF), and Gradient-Boosted Decision Trees (GBDTs), have been widely used for data fusion tasks due to their robustness and interpretability [96], [97], [98]. These methods remain popular, especially when dealing with structured or tabular data.

However, recent advances in DL have led to the development of more complex architectures that are increasingly adopted for data fusion. Convolutional Neural Networks (CNNs), Long Short-Term Memory (LSTM) networks, and Transformer-based models have gained significant attention in the remote sensing-based data fusion [99]. For instance, studies like [100], [101] have demonstrated that Transformer-based fusion methods outperform traditional ML approaches in tasks such as crop yield prediction and soil parameter estimation. These findings underscore the growing importance and potential of DL techniques, particularly in handling large, multi-modal datasets commonly encountered in remote sensing applications.

### C. AN ACTIONABLE ROADMAP FOR DATA FUSION IMPLEMENTATION IN PRECISION AGRICULTURE

To facilitate the practical application of data fusion (DF) in agricultural remote sensing, we propose an actionable, stakeholder-driven roadmap consisting of 4 sequential phases. This roadmap outlines real-world implementation steps, highlights responsible actors, discusses trade-offs, and addresses domain-specific challenges such as spatial heterogeneity, limited ground truth, and computational constraints. Figure 13 provides an overview of this roadmap, tailored for deployment by agricultural researchers and agri-tech developers.

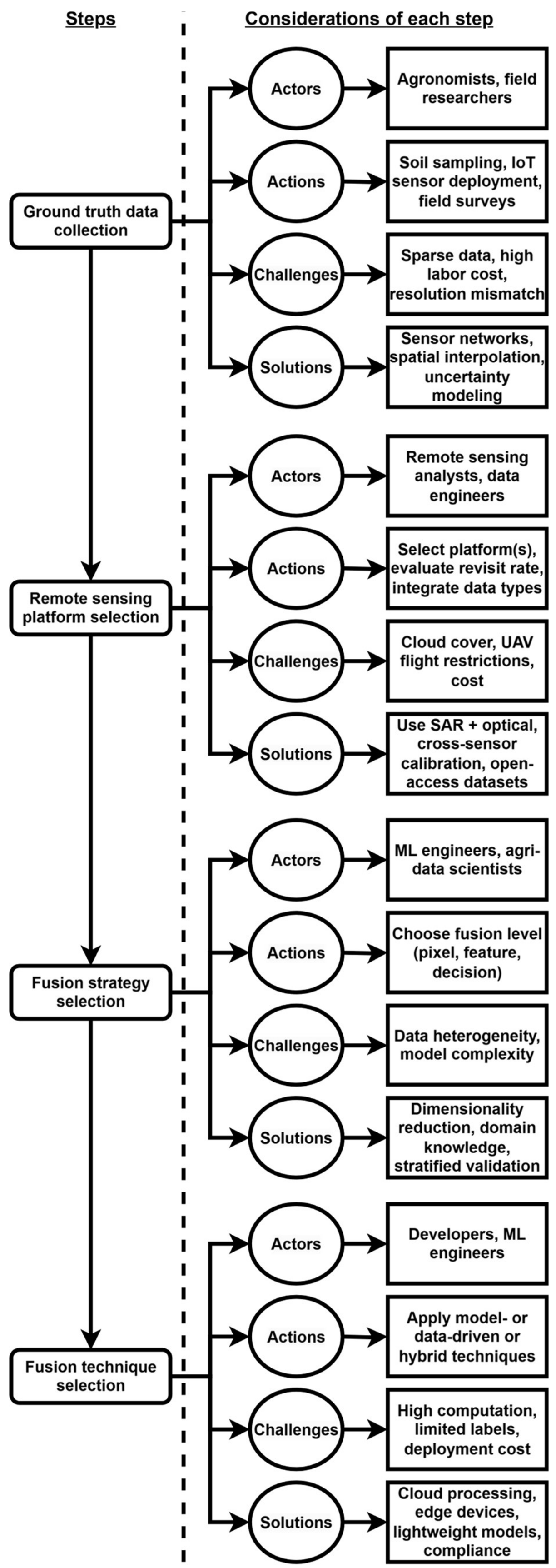

**FIGURE 13.** An actionable roadmap for real-world data fusion implementation in precision agriculture.

### 1) GROUND-TRUTH DATA SELECTION AND QUALITY ASSESSMENT

Establishing a reliable ground-truth dataset is foundational for the development and validation of remote sensing-based data fusion models in agriculture. This task typically involves agronomists, field technicians, researchers, and farm managers who are responsible for the accurate and consistent collection of field measurements. High-quality data are best obtained using calibrated instruments such as TDR sensors for soil moisture or portable spectrometers for vegetation properties [102], [103]. To ensure compatibility with remote sensing inputs, it is essential that field sampling be temporally aligned and spatially co-located.

However, practical challenges such as high labor costs, limited field accessibility, and potential equipment failures can constrain data collection. These issues are often exacerbated by spatial misalignments or data gaps due to weather disruptions. To address these limitations, automated or semi-automated systems based on IoT sensor networks can enhance spatial coverage and reduce manual effort. Additionally, spatial interpolation techniques such as kriging [104] and uncertainty quantification methods such as Monte Carlo simulations [105] or bootstrap analyses [106] can improve the robustness and interpretability of the ground-truth data. Complementing these efforts, crowdsourced or farmer-contributed data platforms [107] can further expand coverage and increase data diversity, especially in resource-constrained regions.

### 2) REMOTE SENSING PLATFORM SELECTION AND SENSOR SYNERGY OPTIMIZATION

The choice of remote sensing platforms and sensors must be closely aligned with the application's objectives, the spatial and temporal resolution requirements and resource constraints. Satellite platforms are particularly effective for broad-scale and long-term monitoring, while UAVs offer high-resolution imagery suited to field-level applications. Airborne platforms serve as an intermediate option, providing regional-scale coverage with customizable configurations [108]. To enhance observation capabilities, combining sensors with complementary characteristics, such as the high temporal frequency of Sentinel-2 with the fine spatial resolution of PlanetScope, can offer synergistic benefits [109].

Nevertheless, several operational constraints exist. Cloud cover, especially in tropical regions, often limits the usability of optical satellite imagery [110]. UAV-based data acquisition is restricted by local aviation regulations, limited flight endurance, and adverse weather conditions [111]. These limitations can be mitigated by integrating synthetic aperture radar (SAR) data, such as from Sentinel-1, which is unaffected by clouds and light conditions [112]. Cross-sensor calibration techniques like histogram matching or radiometric correction are also critical for harmonizing datasets [113]. To reduce the financial burden, especially for smallholder farming systems, open-access sources such as

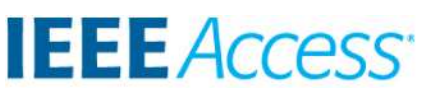

Copernicus [114] or Landsat [115] provide a viable alternative to commercial imagery.

#### 3) DATA FUSION MODE SELECTION BASED ON APPLICATION-SPECIFIC NEEDS

The success of data fusion in agricultural remote sensing heavily depends on choosing the appropriate level of integration, including pixel, feature, or decision level, based on the characteristics of the input data and the target application. Pixel-level fusion is well-suited for fine-grained tasks such as detecting early plant stress or disease [116], where maintaining spatial detail is critical. Feature-level fusion is advantageous when synthesizing heterogeneous data types, such as combining spectral vegetation indices with SAR-derived metrics for comprehensive crop or soil condition assessments [117]. Decision-level fusion becomes essential when integrating independently derived predictions, such as combining outputs from satellite models and IoT-based in-field alerts [118].

However, implementing these fusion strategies requires careful consideration of data compatibility, computational complexity, and noise sensitivity. Large-scale models, particularly those involving high-dimensional inputs, are prone to overfitting and may require substantial computational resources. To address these challenges, dimensionality reduction techniques such as principal component analysis (PCA) [119] or t-distributed stochastic neighbour embedding (t-SNE) [120] can simplify model inputs. Incorporating domain-specific knowledge, such as crop phenological stages, into feature engineering can also enhance model relevance [121]. Moreover, stratified cross-validation based on crop type, soil class, or agroecological zone ensures generalizability and robustness across diverse agricultural contexts [122].

#### 4) ADVANCED DATA FUSION TECHNIQUES AND COMPUTATIONAL CONSIDERATIONS

As fusion methods transition from research prototypes to operational tools, the selection of advanced yet practical techniques becomes paramount. Transformer-based architectures and hybrid models such as CNN-LSTM and Vision Transformers [123] offer powerful capabilities for capturing spatial and temporal dependencies in multimodal datasets. In scenarios with limited labelled data, self-supervised learning (SSL) provides an effective mechanism to pre-train models and reduce dependency on annotated samples [124]. For scalability, cloud-based platforms like Google Earth Engine [125] and AWS SageMaker [126] offer robust infrastructure to handle large-scale agricultural datasets and model deployment.

Despite their advantages, deep learning models often entail high computational costs and energy consumption, posing barriers to adoption in resource-limited environments. This necessitates the development of lightweight architectures, along with model compression techniques such as pruning and quantization. For near real-time applications, especially in drone-based or edge-deployed scenarios, embedded computing solutions such as NVIDIA Jetson [127] platforms can enable on-site inference with minimal latency. Finally, practical deployment must also consider regulatory compliance, including data privacy protection and adherence to UAV operation laws, to ensure ethical and lawful integration into farming workflows.

## VII. CONCLUSION, LIMITATIONS AND FUTURE TRENDS

This review highlights the advancements in Transformer-based data fusion techniques in agricultural remote sensing. The integration of Transformers into remote sensing workflows has led to improvements in both accuracy and efficiency across various applications, including soil moisture prediction, soil element interpretation, and land-use/land-cover classification. These techniques have shown remarkable success in addressing the challenges of integrating multi-source data, with Transformers demonstrating a strong ability to model complex relationships and capture spatial, spectral, and temporal dependencies. As a result, Transformer-based models have become invaluable tools for precision agriculture, providing enhanced decision-making capabilities in crop monitoring, yield prediction, and environmental management.

The analysis of the literature indicates that Transformer-based methods frequently outperform traditional machine learning and deep learning approaches in data fusion tasks. Their ability to capture global contextual information, along with robust feature learning, has contributed to substantial improvements in model performance.

Despite the success of Transformer-based data fusion methods, several challenges remain as follows:

- **High computational cost**: Self-attention mechanisms scale quadratically with input size, resulting in high memory and processing demands, especially when applied to high-resolution or multi-temporal satellite imagery. This limits the deployment of Transformers in operational agricultural systems that require low-latency and resource-efficient processing.
- **Spatial misalignment and heterogeneity**: Multisensor remote sensing data often suffer from spatial misalignments due to differences in spatial resolution, revisit cycles, or sensor characteristics. Transformer models, although powerful, typically assume well-aligned input. Without robust pre-processing or architecture-level corrections, such misalignments can lead to feature confusion and degraded model performance.
- **Ground truth noise and label uncertainty**: Many agricultural datasets rely on in-situ measurements or crowd-sourced labels, which may introduce inconsistencies or noise. Transformer models, being highly data-driven, can overfit to noisy labels unless strategies such as uncertainty modelling, noise-robust loss functions, or ensemble learning are adopted.
- **Domain generalization and transferability**:

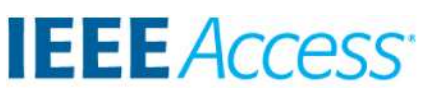

Transformer models trained on specific crops, regions, or seasons may not generalize well to unseen domains due to varying environmental conditions, farming practices, and sensor characteristics. Few-shot adaptation, domain adversarial learning, and meta-learning approaches are needed to enable broader applicability.

- **Dependence on large labelled datasets**: Transformers require large volumes of high-quality labelled data for training. However, in many precision agriculture applications, such data are scarce or expensive to collect. This limitation motivates the adoption of self-supervised, weakly supervised, or transfer learning methods tailored for remote sensing.
- **Limited explainability and interpretability**: While attention maps provide a form of insight into model focus, they do not fully capture causal relationships or actionable reasoning. Practical explainability tools such as SHAP (Shapley Additive Explanations), integrated gradients, and attention-based saliency maps should be integrated into model development pipelines to enable transparent decision-making for agricultural stakeholders.

Looking forward, continued research into Transformer-based data fusion techniques is essential for optimizing agricultural practices and enhancing environmental sustainability. Future studies should focus on improving the generalization of these models, addressing challenges related to data availability and quality, and optimizing computational efficiency. Moreover, exploring the integration of Transformers with other emerging technologies, such as generative models and hybrid frameworks, could lead to even greater advancements in data fusion for agriculture.

Overall, Transformer-based methods offer significant promise for transforming agricultural remote sensing by improving the accuracy, scalability, and robustness of data fusion models. Their application in precision agriculture not only holds the potential to improve productivity and resource efficiency but also contributes to addressing global challenges such as food security and environmental sustainability. As the field continues to evolve, these techniques will play a crucial role in shaping the future of smart agriculture and advancing sustainable farming practices worldwide.

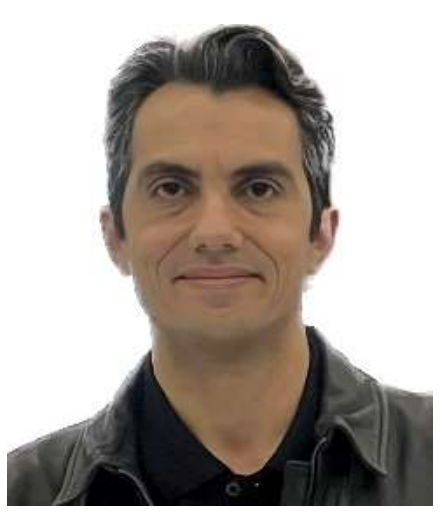

**MAHDI SAKI** received his PhD from the School of Electrical and Data Engineering at the University of Technology Sydney in 2020. His doctoral research focused on big data classification in rail networks. Since completing his PhD, he has made significant contributions to various data-driven industrial and research projects, developing several end-to-end AI/ML pipelines spanning diverse sectors such as transportation, IoT, telecom, and agriculture. He is currently working as an applied scientist in the School of Electrical and Data Engineering at the University of Technology Sydney, where he applies AI/ML techniques in remote sensing-based data fusion in precision agriculture.

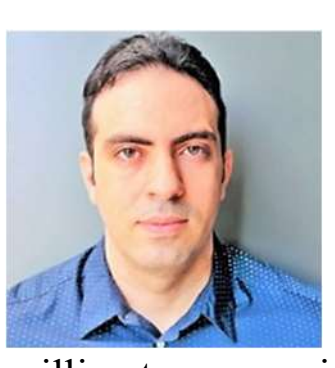

**RASOOL KESHAVARZ** (Senior Member, IEEE) is an accomplished telecommunications engineer with a Ph.D. in Telecommunications Engineering from Amirkabir University of Technology, Tehran, Iran—one of the top three universities in the country. With over 17 years of experience in both industry and academia, he specializes in RF, microwave, and millimeter-wave circuits, radar systems, sensors, and antenna design.

Dr. Keshavarz is currently a Senior Research Fellow and Principal Systems Engineer at the RF and Communication Technologies (RFCT) Lab at the University of Technology Sydney (UTS), Australia, where he plays a key role in leading research initiatives in wireless communication, radar, and sensing technologies. He serves as the technical lead and one of the chief investigators for the project "Sustainable Sensing, Enhanced Connectivity, and Data Analytics for Precision Urban and Rural Agriculture," a collaboration with NTT (Nippon Telegraph and Telephone Corporation) and Food Agility CRC (Cooperative Research Centre). He has also contributed to the successful completion of projects with Zetifi, focusing on "Multi-service, Compact, and Low-cost RF and Antenna Systems for Enhancing Rural and Agricultural Connectivity" and "Farm-wide Wi-Fi: Next Generation RF and Antenna Technologies."

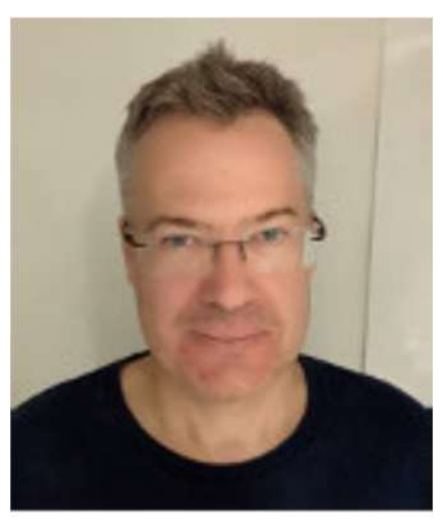

**DANIEL FRANKLIN** (Member, IEEE) received the Bachelor of Engineering degree (Hons.) in electrical from the University of Wollongong, in 1999, and the Ph.D. degree in telecommunications engineering, ‘‘Enhancements to channel Models, DMT Modulation and Coding for Channels Subject to Impulsive Noise,’’ from the University of Wollongong, in 2007. He is currently a Senior Lecturer with the School of Electrical and Data Engineering, University of Technology Sydney. His current research and commercial interests are split between radiation engineering, including positron emission tomography, computed tomography, radiotherapy, radiation dosimetry, pharmacokinetic modeling, and telecommunications engineering, including security applications and network protocols, multimedia, image and signal processing, wired and wireless communication systems, and communications protocols.

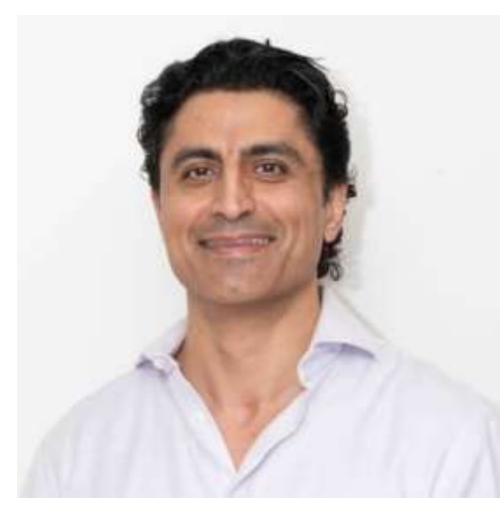

**MEHRAN ABOLHASAN** (Senior Member, IEEE) received the B.E. degree in computer engineering and the Ph.D. degree in telecommunications from the University of Wollongong, in 1999 and 2003, respectively. He has over 20 years of experience in research and development and serving in various research leadership roles. Some of these previous roles include serving as the Director of research programs for the Faculty of Engineering and IT, the Deputy Head of the School for Research with the School of Electrical and Data Engineering in University of Technology Sydney. He is currently the Leader of the Intelligent Networks and Applications Laboratory in Faculty of Engineering and IT, University of Technology Sydney. He has authored over 200 international publications and through his sustained track record of industry engagement, creation of new research ideas and directions, he has successfully secured several R&D projects worth over eight million dollars. This includes external grants such as ARC, CRC, Industry and Government funded research projects. His current research interests include 5G/6G wireless networks, Internet of Things, Cybersecurity, Software Defined Networking, Intelligent Transportation Systems (ITS), Urban Greening and Digital Agriculture.

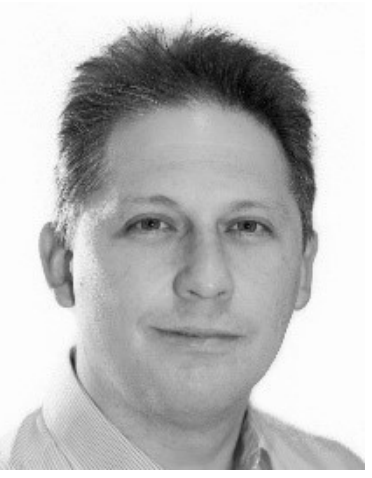

**JUSTIN LIPMAN** (IEEE Senior Member ’12) fosters innovation in connected technologies and thrives at the intersection of academia and industry. As both Industry Associate Professor and Director of the Cyber Digital Centre (CDC) at the University of Technology Sydney, he leverages over 12 years of expertise leading R&D at Intel and Alcatel. Previously at UTS, he was Director of the RF and Communications Technologies Lab and served as Deputy Chief Scientist of the Food Agility Cooperative Research Center. He has secured over $35M in research funding, driving innovation across cybersecurity, RF, IoT, digital agriculture, smart cities, and data privacy. He actively shapes future connected systems through standards development. As a dedicated bridge builder, he fuels impactful industry collaborations, translating research into real-world solutions and positioning himself at the forefront of innovation. He holds 24 U.S. patents and has published over 100 peer-reviewed articles in top-tier conferences and journals.

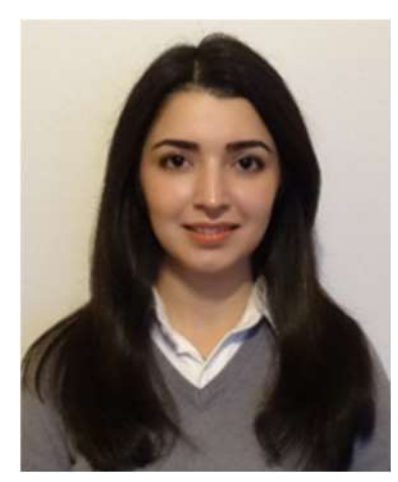

**NEGIN SHARIATI** (Senior Member IEEE) is an Associate Professor at the University of Technology Sydney, Australia. She established the state-of-the-art RF and Communication Technologies (RFCT) research laboratory in 2018, where she is currently the Founding Director, leading R&D in circuits and systems and wireless communication technologies. Dr. Shariati led a 3-year R&D program, "Sensing Innovations Constellation," at the Food Agility Cooperative Research Centre (CRC), which focused on three key interrelated streams in digital agriculture: sensing, energy, and connectivity. This collaborative platform enabled multidisciplinary R&D with industry partners, addressing large-scale, real-world challenges in agricultural technologies. She also served as the Director of Women in Engineering and IT at UTS, driving positive changes in equity and diversity in STEM. Since 2018, she has held a joint academic appointment at Hokkaido University, engaging in research and teaching activities in Japan. Dr. Shariati has secured more than six million dollars in research funding across multiple ARC, CRC, industry, and government-funded research projects, where she has taken the lead Chief Investigator (CI) role and contributed as a CI team member. She leads a 3-year multidisciplinary R&D program 'Sustainable Sensing, Enhanced Connectivity, and Data Analytics for Precision Urban and Rural Agriculture' in collaboration with NTT and Food Agility CRC. She completed her PhD in Electrical-Electronics and Communication Technologies at RMIT University, Australia, in 2016. Prior to that, she worked in industry as an Electrical-Electronics Engineer 2009-2012.